\pdfoutput=1

\documentclass[11pt]{article}

\usepackage[preprint]{acl}

\usepackage{times}
\usepackage{latexsym}

\usepackage[T1]{fontenc}

\usepackage[utf8]{inputenc}

\usepackage{microtype}

\usepackage{inconsolata}

\usepackage{graphicx}
\usepackage{amsmath}
\usepackage{amsfonts}
\usepackage{dsfont}
\usepackage{booktabs}
\usepackage{multirow}
\usepackage{float}
\usepackage{afterpage}
\usepackage{placeins}
\usepackage{enumitem}

\title{SMARTe: Slot-based Method for Accountable Relational Triple extraction}

\author{
\textbf{
Xue Wen Tan\textsuperscript{\rm 1}~~
Stanley Kok\textsuperscript{\rm 2}~~} \\
\textsuperscript{\rm 1}Asian Institute of Digital Finance, National University of Singapore \\
\textsuperscript{\rm 2}School of Computing, National University of Singapore\\
{\tt xuewen@u.nus.edu},~~{\tt skok@comp.nus.edu.sg}\\
}

\begin{document}
\maketitle
\begin{abstract}
Relational Triple Extraction (RTE) is a fundamental task in Natural Language Processing (NLP). However, prior research has primarily focused on optimizing model performance, with limited efforts to understand the internal mechanisms driving these models. Many existing methods rely on complex preprocessing to induce specific interactions, often resulting in opaque systems that may not fully align with their theoretical foundations. To address these limitations, we propose SMARTe: a Slot-based Method for Accountable Relational Triple extraction. SMARTe introduces intrinsic interpretability through a slot attention mechanism and frames the task as a set prediction problem. Slot attention consolidates relevant information into distinct slots, ensuring all predictions can be explicitly traced to learned slot representations and the tokens contributing to each predicted relational triple. While emphasizing interpretability, SMARTe achieves performance comparable to state-of-the-art models. Evaluations on the NYT and WebNLG datasets demonstrate that adding interpretability does not compromise performance. Furthermore, we conducted qualitative assessments to showcase the explanations provided by SMARTe, using attention heatmaps that map to their respective tokens. We conclude with a discussion of our findings and propose directions for future research.
\end{abstract}

\section{Introduction}
Relational Triple Extraction (RTE) is a well-established and widely studied task in Natural Language Processing (NLP). Its primary objective is to automatically extract structured information such as names, dates, and relationships from unstructured text, thereby enhancing data organization and accessibility \cite{nayak2021deep}. These extracted relationships are represented as relational triples consisting of \texttt{(Subject, Relation, Object)}.Such structured representations underpin a range of downstream applications, including knowledge graph construction, question answering, and information retrieval. RTE is often confused with Joint Entity and Relation Extraction (JERE) \cite{zhang2017end, gupta2016table, miwa2014modeling}, or the terms are used interchangeably by some scholars \cite{sui2023joint, li2021tdeer}. However, there is a key distinction: JERE addresses both entity identification and relation extraction (i.e., ACE04, ACE05 and CoNLL04 datasets) \cite{roth2004linear}, which require annotation of all entities in a text, regardless whether they participate in a relationship. In contrast, RTE focuses exclusively on entities that are part of a relationship, specifically a \textbf{\textit{set of annotated relational triples}}, hence the two tasks are fundamentally different. In this paper, we focus \textbf{solely on RTE}.

\begin{figure}[hbt!]
\centering
  \includegraphics[width=0.80\linewidth]{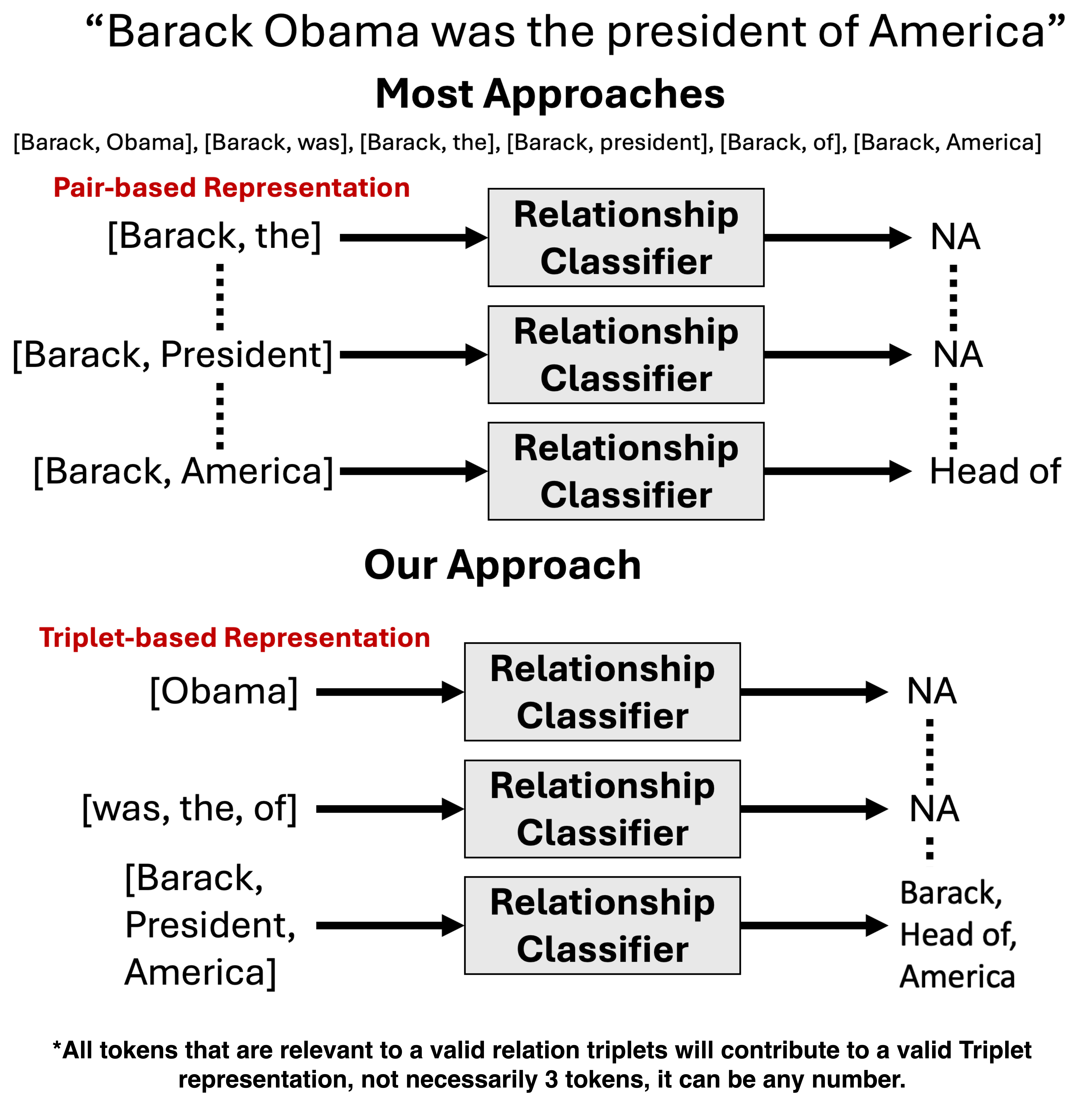}
  \caption{Most Approaches vs Our Approach.}
  \label{fig:motivation}
\end{figure}

\begin{figure*}[hbt!]
\centering
  \includegraphics[width=0.80\linewidth]{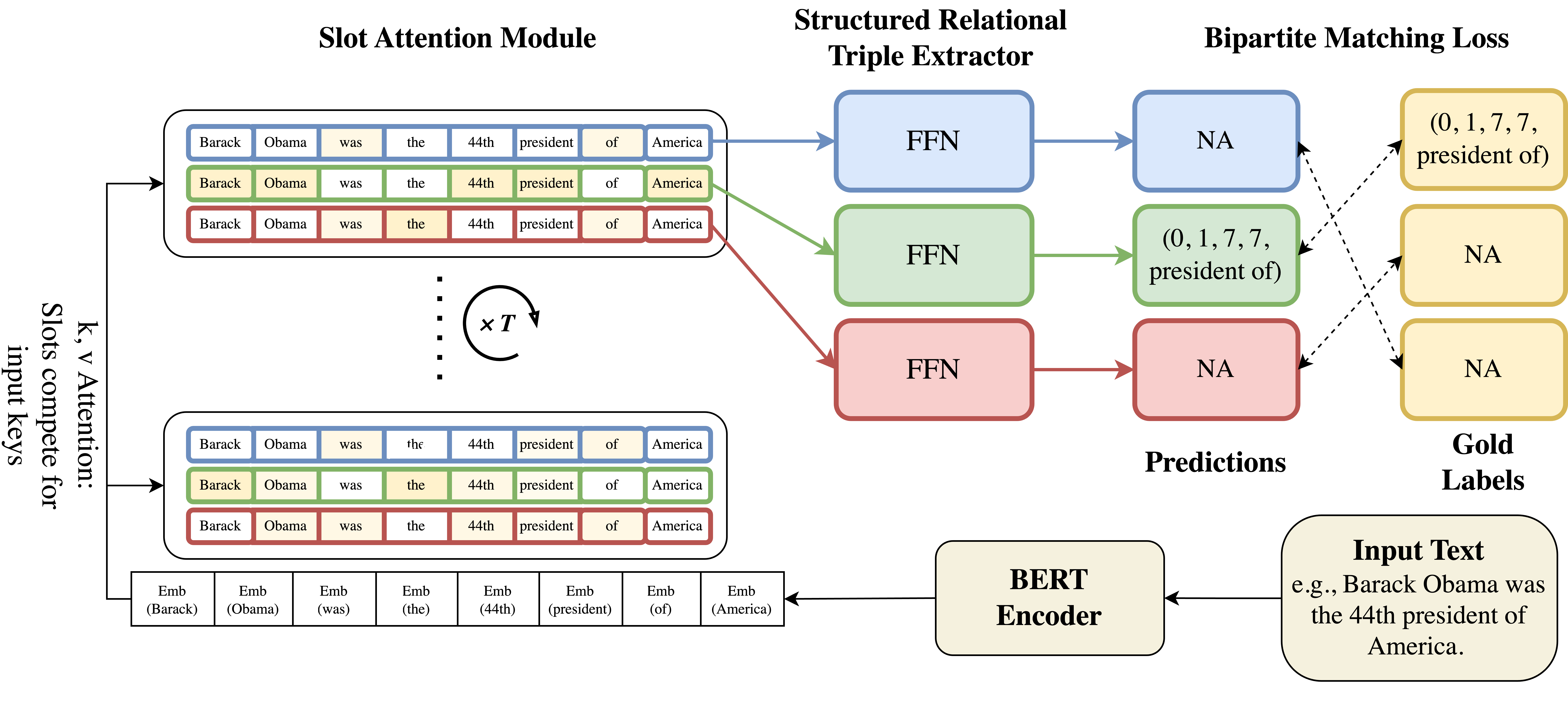}
  \caption{Architecture Diagram of our SMARTe model. The Slot Attention module iteratively refines the slots over T iterations to obtain a final representation, from which relational triples are extracted.}
  \label{fig:architecturediagram}
\end{figure*}

To motivate our work, lets take a look at an example seen in Figure \ref{fig:motivation}.
Most approaches rely on fine-tuned BERT embeddings to form \textit{pair representations} by permuting all token pairs and classifying their relationships. However, this methodology makes it difficult to trace which specific contextual information each pair has absorbed, for instance, whether "Barack" or "America" has incorporated the "president" context. Despite achieving superior F1 performance, these pair-based approaches lack interpretability, obscuring how models arrive at their predictions. In contrast, our approach uses \textit{triplet-based representation}, which absorbs all relevant tokens to directly form relational triplets. This enables a more direct and interpretable encoding of relationships. We observed that most existing methods have excellent performance but with limited interpretability, which highlights a significant gap in current research. In this study, we introduce \textbf{SMARTe}\footnote{We will release our implementation after acceptance.}: a \textbf{S}lot-based \textbf{M}ethod for \textbf{A}ccountable \textbf{R}elational \textbf{T}riple Extraction, a transparent architecture designed to address interpretability in RTE. SMARTe incorporates intrinsic interpretability through the use of a slot attention mechanism. This approach ensures that every prediction from SMARTe can be explicitly traced back to its learned slot representations and the specific tokens contributing to each predicted relational triple as shown in Appendix \ref{app:A}. This results in informed and transparent outputs, embodying the \textit{\textbf{A}ccountability, which directly equates to explainability}, in our model. Since the learned slots do not follow any inherent order, the RTE task is framed as a \textit{set prediction} problem. 

Although we illustrate our approach in the context of RTE, it is \textit{\textbf{broadly applicable to other NLP tasks characterized by set-based target structures}}. Interpretability becomes especially valuable in high-stakes scenarios where transparency and trust are paramount \cite{danilevsky2020survey}. As stakeholders increasingly demand clear and interpretable explanations for AI-driven decisions, our approach effectively addresses these critical requirements. We introduce \textbf{SMARTe}, a \textbf{S}lot-based \textbf{M}ethod for \textbf{A}ccountable \textbf{R}elational \textbf{T}riple \textbf{e}xtraction. Our contributions are as follows:
\begin{itemize}[leftmargin=*, itemsep=1pt]
\item To the best of our knowledge, the proposed SMARTe framework is the first to introduce interpretability to relational triple extraction tasks by adapting the slot attention mechanism, originally developed for unsupervised learning in computer vision.
\item We conduct extensive experiments on two widely used datasets and demonstrate that our model achieves performance comparable to current state-of-the-art systems, all while offering interpretability.
\item We provide a qualitative assessment and demonstrate how slot attention facilitates explanations which allow users to understand the model's reasoning behind its predictions.
\item We report zero-shot performance on RTE tasks using recent LLMs for future benchmarks.
\end{itemize}
\section{Related Work}
Recent advancements in relation triplet extraction (RTE) have investigated various architectures to better capture interactions between entities and relations. We group these approaches into three main categories: sequence-to-sequence (seq2seq) Methods, Tagging-Based Methods, and Pairwise-Based Methods.

\textbf{Seq2seq methods} treat triples as token sequences, leveraging an encoder-decoder framework akin to machine translation. CopyRE \cite{zeng2018extracting} uses a copy mechanism to generate relations and entities but struggles with multi-token entities. CopyMTL \cite{zeng2020copymtl} addresses this by employing a multi-task learning framework. CGT \cite{ye2021contrastive} introduces a generative transformer with contrastive learning to enhance long-term dependency and faithfulness. R-BPtrNet \cite{chen2021jointly} uses a binary pointer network to extract explicit and implicit triples, while SPN \cite{sui2023joint} reframes relational triple extraction as a set prediction problem, utilizing a non-autoregressive decoder with iterative refinement for improved contextual representation.

\textbf{Tagging-Based Methods}, also known as sequence labeling methods, utilize binary tagging sequences to identify the start and end positions of entities and, in some cases, to determine relations. Early approaches, such as NovelTagging \cite{zheng2017joint}, introduced a tagging-based framework that reformulates joint extraction as a tagging problem, enabling direct extraction of entities and their relations. CasRel \cite{wei2019novel} improves on this by first identifying all potential head entities and then applying relation-specific sequence taggers to detect corresponding tail entities. Recent methods, however, are not entirely sequence-based. For instance, BiRTE employs tagging only for entities. It employs a pipeline strategy in which entities are tagged as either subjects or objects, after which all possible subject-object pairs are generated and classified using a biaffine scorer. Similarly, PRGC \cite{zheng2021prgc} includes a component to predict potential relations, constraining subsequent entity recognition to the predicted relation subset. PRGC's strong performance stems from its use of a global correspondence table, which effectively captures interactions between token pairs.

\textbf{Pairwise-Based Methods} focus on enhancing token-pair interaction representations to improve relation classification. These methods eliminate the need to explicitly predict head or tail entities, as token pairs classified as \texttt{NA} (no relationship) are directly discarded. Notably, approaches in this category have achieved state-of-the-art (SOTA) F1 performance. Early works often framed these methods as table-filling approaches. For example, GraphRel \cite{fu2019graphrel} models entity-relation interactions through a relation-weighted Graph Convolutional Network. TPLinker \cite{wang2020tplinker} redefines triple extraction as a token-pair linking task, utilizing a relation-specific handshaking tagging scheme to align boundary tokens of entity pairs. Similarly, PFN \cite{yan2021partition} introduces a partition filter network that integrates task-specific feature generation to simultaneously model entity recognition and relation classification. Modern SOTA techniques further refine these approaches. UniRel \cite{tang2022unirel} employs a unified interaction map to effectively capture token-relation interactions and also incorporating relation-specific token information into the prediction process. Similarly, DirectRel \cite{ijcai2022p605} reformulates relational triple extraction (RTE) as a bipartite graph linking task, focusing on generating head-tail entity pairs.

\section{Approach}
The architecture of our SMARTe model is depicted in Figure \ref{fig:architecturediagram}, consisting of three primary components: an encoder, a slot attention module and a structured relational triple extractor.

\subsection{Encoder}
The encoder transforms the input text into dense, contextual representations that capture the semantic and syntactic information necessary for downstream processing. In our implementation, we utilize pre-trained transformer-based models BERT-Base-Cased \cite{devlin-etal-2019-bert}. The process begins with tokenizing the input text into subword units, which are then fed into the transformer. The output is a sequence of contextualized embeddings, each corresponding to a token in the input text. Given an input text sequence $X={x_1,x_2, \ldots ,x_n}$, where $x_i$ represents the $i$-th token, the encoder transforms this sequence into a sequence of contextualized embeddings:
\begin{equation}
    \mathbf{H} = \{\mathbf{h}_1, \mathbf{h}_2, \ldots, \mathbf{h}_n\} = \text{Encoder}(X) .
\end{equation}
Here, $\mathbf{H} \in \mathbb{R}^{n\times d}$, where $n$ is the length of a sequence of contextualized embeddings (including \texttt{[CLS]} and \texttt{[SEP]}, two special start and end markers), with each $\mathbf{h}_i$ corresponding to a token in the input text, and $d$ is the embedding size produced by the encoder for each token.

\subsection{Slot Attention Module}
Slot Attention is a neural network module designed for object-centric representation learning \cite{locatello2020object}, enabling a model to decompose a scene into distinct entities or objects. It uses iterative attention mechanisms to map input data (like image features) to a fixed number of slots, which are learnable feature vectors representing objects or parts of the input. This approach is highly effective in unsupervised learning settings, as it disentangles objects without requiring explicit annotations. The general form of Slot Attention is as follow:
\begin{flalign}
& \mathbf{Q}^{(l)} = \mathbf{Z}^{(l)} \mathbf{W}_Q,
\mathbf{K} = \mathbf{X} \mathbf{W}_K,
\mathbf{V} = \mathbf{X} \mathbf{W}_V & \\
& \mathbf{A}^{(l)} = \text{normalize}(\text{softmax}(\mathbf{Q}^{(l)} \mathbf{K}^\top)) & \label{eq:softmax} \\
& \mathbf{Z}^{(l+1)} = \text{GRU}(\mathbf{Z}^{(l)}, \mathbf{A}^{(l)} \mathbf{V}). &
\end{flalign}
At iteration \( l \), the current slots $ \mathbf{Z}^{(l)}$ are transformed into query vectors \( \mathbf{Q}^{(l)} \) using a learnable weight matrix \( \mathbf{W}_Q \), while the input features \( \mathbf{X} \), derived from the sequence of contextualized embeddings from the encoder \( \mathbf{H} \), are projected into key \( \mathbf{K} \) and value \( \mathbf{V} \) vectors using \( \mathbf{W}_K \) and \( \mathbf{W}_V \), respectively. Attention weights \( \mathbf{A}^{(l)} \) are computed as the dot product between \( \mathbf{Q}^{(l)} \) and \( \mathbf{K}^\top \), followed by softmax to produce a probability distribution that ensures each slot attends to specific parts of the input. The slots are updated using a Gated Recurrent Unit (GRU) \cite{chung2014empirical}, combining the previous slot representation \( \mathbf{Z}^{(l)} \) with the weighted aggregation of values \( \mathbf{A}^{(l)} \mathbf{V} \). This iterative process refines the slots, allowing the model to disentangle objects or entities in the input data. After completing the iterative process, the output is a refined slots \( \mathbf{Z} \in \mathbb{R}^{k \times d} \), where \( k \) is the number of slots and \( d \) is the dimensionality of each slot.

However, the softmax function in slot attention can be too restrictive for relational triple extraction tasks, particularly when certain tokens, such as ``Barack Obama,'' are involved in multiple triples and need to be associated with multiple slots. For instance, in the triples \texttt{(Barack Obama, Born In, Hawaii)}, \texttt{(Barack Obama, President Of, United States)}, and \texttt{(Barack Obama, Married To, Michelle Obama)}, the token \texttt{Barack Obama} serves as the subject in each case. Softmax enforces a normalization constraint where attention scores must sum to 1, causing the token's contribution to be distributed across slots. This can dilute its impact and make it difficult for the model to maintain consistent associations across triples. To address this limitation, our SMARTe model employ the optimal transport variant outlined in \citet{pmlr-v202-zhang23ba}, which is more relaxed and provides a flexible framework for assigning tokens to slots while preserving their relevance across multiple contexts. Specifically, this involves replacing equation \ref{eq:softmax} with the optimal transport algorithm \cite{villani2009optimal}, as follows:
\begin{flalign}
& \mathbf{C}' = \frac{\mathbf{Q} \cdot \mathbf{K}}{\|\mathbf{Q}\| \|\mathbf{K}\|} & \label{eq:cossim}\\
& \text{MESH}(\mathbf{C}) = \mathop{\arg \min}\limits_{\mathbf{C}' \in \mathcal{V}(\mathbf{C})} H(\text{sinkhorn}(\mathbf{C}')) & \label{eq:mesh} \\
& \mathbf{A} = \text{sinkhorn}(\text{MESH}(\mathbf{C})). & \label{eq:meshA}
\end{flalign}
In equation \ref{eq:cossim}, the initial transport cost \(\mathbf{C}'\) is computed as the cosine similarity between \(\mathbf{Q}\) and \(\mathbf{K}\). Next, in equation \ref{eq:mesh}, \(\text{MESH}(\mathbf{C})\) selects, from the set \(\mathcal{V}(\mathbf{C})\), the candidate matrix \(\mathbf{C}'\) that minimizes the entropy of the initial cost function, \(H\bigl(\text{sinkhorn}(\mathbf{C}')\bigr)\). Here, \(\text{sinkhorn}(\cdot)\) iteratively normalizes the row and column sums to produce a doubly stochastic matrix. \(\text{MESH}(\cdot)\) corresponds to Minimizing the Entropy of Sinkhorn. Finally, in equation \ref{eq:meshA}, the chosen matrix \(\text{MESH}(\mathbf{C})\) is passed once more through the Sinkhorn operator to yield the final attention matrix \(\mathbf{A}\). This “transport plan” preserves row- and column-stochastic constraints, providing a refined set of attention weights. For more details, we refer readers to the work of \citet{pmlr-v202-zhang23ba}. We have also provided the results and analysis for the softmax variant for reference.

\subsection{Structured Relational Triple Extractor}
In the original Slot Attention paper \cite{locatello2020object}, the approach was designed for fixed-size images. However, when applied to text, variable input lengths make direct coordinate prediction more challenging. To address this, we reformulate coordinate prediction as a \textit{one-hot sequence labeling task} to identify the start and end positions. We perform matrix multiplication between the slot attention outputs and the sequence tokens to encode information from the slot attention representation into each token. A feedforward neural network with linear layers predicts the indices for entity head-tail pairs, along with their corresponding relationships, in a unified manner. Specifically, it identifies five key components: \texttt{[subject-start (ss), subject-end (se), object-start (os), object-end (oe), and relationship (rs)]}, which collectively form a structured relational triple:
\begin{flalign}
    & \mathbf{P}_{i}^{\text{ss}} = \sigma(\mathbf{V}^{\top}_{1}\text{tanh}(\mathbf{W}_{1}\mathbf{Z}_{i} + \mathbf{W}_{2}\mathbf{H})) & \\
    & \mathbf{P}_{i}^{\text{se}} = \sigma(\mathbf{V}^{\top}_{2}\text{tanh}(\mathbf{W}_{3}\mathbf{Z}_{i} + \mathbf{W}_{4}\mathbf{H})) & \\
    & \mathbf{P}_{i}^{\text{os}} = \sigma(\mathbf{V}^{\top}_{3}\text{tanh}(\mathbf{W}_{5}\mathbf{Z}_{i} + \mathbf{W}_{6}\mathbf{H})) & \\
    & \mathbf{P}_{i}^{\text{oe}} = \sigma(\mathbf{V}^{\top}_{4}\text{tanh}(\mathbf{W}_{7}\mathbf{Z}_{i} + \mathbf{W}_{8}\mathbf{H})) & \\
    & \mathbf{P}_{i}^{\text{rs}} = \sigma(\mathbf{W}_{t}\mathbf{Z}_{i}). &
\end{flalign}
$\mathbf{P}_{i}^{*}$ refers to the prediction for each component at the $i^{th}$ slot, with $\mathbf{Z}_{i}$ representing the $i^{th}$ slot embedding; the matrices $\mathbf{W}_{t} \in \mathbb{R}^{t \times d}$, $[\mathbf{W}_{i}]_{i=1}^{8} \in \mathbb{R}^{d \times d}$, where $t$ represents the total number of relation types in the RTE task and $[\mathbf{V}_{i}]_{i=1}^{4} \in \mathbb{R}^{d}$ are learnable parameters. The function $\sigma$ refers to the \texttt{softmax} operation and $\mathbf{H}$ represents the sequence of contextualized embeddings.

Since the slots inherently lack ordering, directly comparing predictions with the ground truth during model training presents a challenge. To overcome this, the predictions must first be aligned with the ground truth. This alignment is achieved using the Hungarian matching algorithm \cite{kuhn1955hungarian}. After optimally matching the predictions to the ground truth, cross-entropy loss is applied to each component to train the model:
\begin{flalign}
& \mathcal{L} = \sum^{k}_{i=1} \{ -\log \mathbf{P}_{\pi^{*}}^{\text{rs}}(\mathbf{T}_{i}^{\text{rs}}) + \mathds{1}_{\{\mathbf{T}_{i}^{\text{rs}} \neq \texttt{NA}\}} \nonumber & \\
& \quad\quad [-\log \mathbf{P}_{\pi^{*}}^{\text{ss}}(\mathbf{T}_{i}^{\text{ss}})-\log \mathbf{P}_{\pi^{*}}^{\text{se}}(\mathbf{T}_{i}^{\text{se}}) \nonumber & \\
& \quad\quad -\log \mathbf{P}_{\pi^{*}}^{\text{os}}(\mathbf{T}_{i}^{\text{os}})-\log \mathbf{P}_{\pi^{*}}^{\text{oe}}(\mathbf{T}_{i}^{\text{oe}})]\} & 
\end{flalign}

\noindent where $\pi^{*}$ represents the optimal assignment that minimizes the total pairwise matching cost and $\mathbf{P}^{*}_{\pi(i)}$ denotes the prediction, $\mathbf{T}^{*}_{i}$ represents the corresponding ground truth target at the $i^{th}$ slot and the indicator function $\mathds{1}_{\{\mathbf{T}_{i}^{\text{rs}} \neq \texttt{NA}\}}$ ensuring only valid relational triples contribute to the loss.

\section{Experimental Setup}
We evaluate our model using two widely recognized benchmark datasets frequently employed in the existing literature on relational triple extraction: the New York Times (NYT) dataset \cite{riedel2010modeling} and the Web Natural Language Generation (WebNLG) dataset \cite{gardent2017creating} (Please note that these are the only two benchmark datasets specifically designed for this task). Both datasets include two variants of annotation: Partial Matching and Exact Matching. In Partial Matching, only the head words of the ground truth entities are annotated, whereas in Exact Matching, the entire span of the entities is annotated. Therefore, Exact Matching offers a more precise and comprehensive representation of the task. It is worth mentioning that most prior studies primarily report results based on Partial Matching, often neglecting Exact Matching. In our study, we present \textbf{results for both matching strategies} to facilitate comparisons in future research. Detailed statistics for these datasets are provided in Table \ref{tab:dataset_statistics} and \ref{tab:overlapping_statistics}. Please refer to Appendix \ref{app:C} for a detailed overview of the key hyperparameters and hardware used in our experiments.

\section{Experimental Results}
\begin{table*}[hbt!]
\centering
\caption{Precision (\%), Recall (\%), and F1-score (\%) of SMARTe and baselines on the NYT dataset (* indicates partial matching, while exact matching is indicated without it). ${\dagger}$ represents the best-performing seed in terms of overall F1 score for reference. $\pm$ represents the standard deviation of the results across 12 runs. $^{\ddagger}$ refers to zero-shot prompting for the LLMs as seen in appendix \ref{app:prompt}.}
\begin{tabular}{lllllll}
\toprule
\multirow{2}{*}{Model} & \multicolumn{3}{c}{NYT*} & \multicolumn{3}{c}{NYT} \\
 & Prec. & Rec. & F1 & Prec. & Rec. & F1 \\
\midrule
CasRel \cite{wei2019novel} & 89.7 & 89.5 & 89.6 & 90.1 & 88.5 & 89.3 \\
TPLinker \cite{wang2020tplinker} & 91.3 & 92.5 & 91.9 & 91.4 & 92.6 & 92.0 \\
CGT \cite{ye2021contrastive} & \textbf{94.7} & 84.2 & 89.1 & - & - & - \\
PRGC \cite{zheng2021prgc} & 93.3 & 91.9 & 92.6 & \underline{93.5} & 91.9 & 92.7 \\
R-BPtrNet \cite{chen2021jointly} & 92.7 & 92.5 & 92.6 & - & - & - \\
BiRTE \cite{ren2022simple} & 92.2 & \underline{93.8} & 93.0 & 91.9 & \textbf{93.7} & \underline{92.8} \\
DirectRel \cite{ijcai2022p605} & \underline{93.7} & 92.8 & \underline{93.2} & \textbf{93.6} & 92.2 & \textbf{92.9} \\
UniRel \cite{tang2022unirel} & 93.5 & \textbf{94.0} & \textbf{93.7} & - & - & - \\
SPN \cite{sui2023joint} & 93.3 & 91.7 & 92.5 & 92.5 & 92.2 & 92.3 \\
Phi4-14B$^{\ddagger}$ \cite{abdin2024phi}  & - & - & - & 3.3 & 6.1 & 4.0 \\
Gemma3-27B$^{\ddagger}$ \cite{gemma_2025} & - & - & - & 6.0 & 12.0 & 7.3 \\
Qwen3-32B$^{\ddagger}$ \cite{qwen3} & - & - & - & 3.7 & 8.2 & 4.7 \\
Llama3.3-70B$^{\ddagger}$ \cite{grattafiori2024llama} & - & - & - & 7.0 & 15.2 & 8.9 \\
GPT 4o-Mini$^{\ddagger}$ \cite{openai2024gpt4technicalreport} & - & - & - & 3.3 & 6.3 & 3.9 \\
GPT 4.1-Mini$^{\ddagger}$ \cite{openai2024gpt4technicalreport} & - & - & - & 7.1 & 11.3 & 7.9 \\
\midrule
SMARTe (Softmax) & 92.2$_{\scalebox{0.4}{\(\pm 0.3\)}}$ & 91.2$_{\scalebox{0.4}{\(\pm 0.4\)}}$ & 91.7$_{\scalebox{0.4}{\(\pm 0.3\)}}$ & 
92.1$_{\scalebox{0.4}{\(\pm 0.2\)}}$ & 
91.7$_{\scalebox{0.4}{\(\pm 0.2\)}}$ & 
91.9$_{\scalebox{0.4}{\(\pm 0.2\)}}$ \\
\textbf{SMARTe (Opt Transport)} & 92.4$_{\scalebox{0.4}{\(\pm 0.3\)}}$ & 92.9$_{\scalebox{0.4}{\(\pm 0.2\)}}$ & 92.6$_{\scalebox{0.4}{\(\pm 0.2\)}}$ & 92.5$_{\scalebox{0.4}{\(\pm 0.2\)}}$ & 93.0$_{\scalebox{0.4}{\(\pm 0.1\)}}$ & 92.7$_{\scalebox{0.4}{\(\pm 0.1\)}}$ \\
\textbf{SMARTe$^{\dagger}$ (Opt Transport)} & 92.5 & 93.3 & 92.9 & 92.7 & \underline{93.1} & \textbf{92.9} \\
\bottomrule
\end{tabular}
\label{tab:nyt_results}
\end{table*}
\begin{table*}[hbt!]
\centering
\caption{Precision (\%), Recall (\%), and F1-score (\%) of SMARTe and baselines on the WebNLG dataset (* indicates partial matching, while exact matching is indicated without it). ${\dagger}$ represents the best-performing seed in terms of overall F1 score for reference. $\pm$ represents the standard deviation of the results across 12 runs. $^{\ddagger}$ refers to zero-shot prompting for the LLMs as seen in appendix \ref{app:prompt}.}
\begin{tabular}{lllllll}
\toprule
\multirow{2}{*}{Model} & \multicolumn{3}{c}{WebNLG*} & \multicolumn{3}{c}{WebNLG} \\
 & Prec. & Rec. & F1 & Prec. & Rec. & F1 \\
\midrule
CasRel \cite{wei2019novel} & 93.4 & 90.1 & 91.8 & - & - & - \\
TPLinker \cite{wang2020tplinker} & 91.7 & 92.0 & 91.9 & 88.9 & 84.5 & 86.7 \\
CGT \cite{ye2021contrastive} & 92.9 & 75.6 & 83.4 & - & - & - \\
PRGC \cite{zheng2021prgc} & 94.0 & 92.1 & 93.0 & 89.9 & 87.2 & 88.5 \\
R-BPtrNet \cite{chen2021jointly} & 93.7 & 92.8 & 93.3 & - & - & - \\
BiRTE \cite{ren2022simple} & 93.2 & 94.0 & 93.6 & 89.0 & \textbf{89.5} & 89.3 \\
DirectRel \cite{ijcai2022p605} & \underline{94.1} & \underline{94.1} & \underline{94.1} & \textbf{91.0} & 89.0 & \textbf{90.0} \\
UniRel \cite{tang2022unirel} & \textbf{94.8} & \textbf{94.6} & \textbf{94.7} & - & - & - \\
SPN \cite{sui2023joint} & 93.1 & 93.6 & 93.4 & - & - & - \\
Phi4-14B$^{\ddagger}$ \cite{abdin2024phi} & - & - & - & 13.5 & 16.4 & 14.3 \\
Gemma3-27B$^{\ddagger}$ \cite{gemma_2025} & - & - & - & 24.6 & 31.4 & 26.6 \\
Qwen3-32B$^{\ddagger}$ \cite{qwen3} & - & - & - & 15.7 & 19.2 & 16.8 \\
Llama3.3-70B$^{\ddagger}$ \cite{grattafiori2024llama} & - & - & - & 25.7 & 31.3 & 27.3 \\
GPT 4o-Mini$^{\ddagger}$ \cite{openai2024gpt4technicalreport} & - & - & - & 24.0 & 28.8 & 25.3 \\
GPT 4.1-Mini$^{\ddagger}$ \cite{openai2024gpt4technicalreport} & - & - & - & 29.6 & 32.8 & 30.1 \\
\midrule
SMARTe (Softmax) & 92.8$_{\scalebox{0.4}{\(\pm 0.4\)}}$ & 92.5$_{\scalebox{0.4}{\(\pm 0.4\)}}$ & 92.7$_{\scalebox{0.4}{\(\pm 0.3\)}}$ & 85.2$_{\scalebox{0.4}{\(\pm 0.2\)}}$ & 83.8$_{\scalebox{0.4}{\(\pm 0.4\)}}$ & 84.5$_{\scalebox{0.4}{\(\pm 0.2\)}}$ \\
\textbf{SMARTe (Opt Transport)} & 93.4$_{\scalebox{0.4}{\(\pm 0.4\)}}$ & 93.4$_{\scalebox{0.4}{\(\pm 0.3\)}}$ & 93.4$_{\scalebox{0.4}{\(\pm 0.3\)}}$ & 86.0$_{\scalebox{0.4}{\(\pm 0.4\)}}$ & 85.6$_{\scalebox{0.4}{\(\pm 0.4\)}}$ & 85.8$_{\scalebox{0.4}{\(\pm 0.3\)}}$ \\
\textbf{SMARTe$^{\dagger}$ (Opt Transport)} & 93.7 & 94.0 & 93.9 & 86.6 & 86.0 & 86.3 \\
\bottomrule
\end{tabular}
\label{tab:webnlg_results}
\end{table*}
\setlength{\tabcolsep}{4pt}
\begin{table*}[ht]
\centering
\caption{F1-score (\%) on sentences with different overlapping patterns and triple numbers for NYT* test set. ${\dagger}$ represents the best-performing seed in terms of overall F1 score for reference. $\pm$ represents the standard deviation of the results across 12 runs.}
\begin{tabular}{@{}lllllllll@{}}
\toprule
Model      & Normal & EPO  & SEO  & N=1  & N=2  & N=3  & N=4  & N$\geq$5 \\
\midrule
CasRel \cite{wei2019novel}     & 87.3   & 92.0 & 91.4 & 88.2  & 90.3  & 91.9  & 94.2 & 83.7 \\
TPLinker \cite{wang2020tplinker}  & 90.1   & 94.0 & 93.4 & 90.0  & 92.8  & 93.1  & 96.1 & 90.0 \\
PRGC \cite{zheng2021prgc} & 91.0   & 94.5 & 94.0 & 91.1  & 93.0  & 93.5  & 95.5 & 93.0 \\
BiRTE \cite{ren2022simple} & 91.4   & 94.2 & \underline{94.7} & \underline{91.5}  & 93.7  & 93.9 & 95.8 & 92.1 \\
DirectRel \cite{ijcai2022p605} & \textbf{91.7}   & 94.8 & 94.6 & \textbf{91.7}  & 94.1  & 93.5  & 96.3 & \underline{92.7} \\
UniRel \cite{tang2022unirel} & \underline{91.6} & \textbf{95.2} & \textbf{95.3} & \underline{91.5} & \textbf{94.3} & \textbf{94.5}  & \textbf{96.6}  & \textbf{94.2} \\
SPN \cite{sui2023joint} & 90.8 & 94.1 & 94.0 & 90.9  & 93.4  & 94.2  & 95.5  & 90.6 \\
\hline
SMARTe (Softmax) & 
90.3$_{\scalebox{0.4}{\(\pm 0.3\)}}$ & 
93.4$_{\scalebox{0.4}{\(\pm 0.4\)}}$ & 
92.9$_{\scalebox{0.4}{\(\pm 0.4\)}}$ & 
90.3$_{\scalebox{0.4}{\(\pm 0.3\)}}$ & 
93.1$_{\scalebox{0.4}{\(\pm 0.4\)}}$ & 
93.3$_{\scalebox{0.4}{\(\pm 0.4\)}}$ & 
95.5$_{\scalebox{0.4}{\(\pm 0.5\)}}$ & 
85.3$_{\scalebox{0.4}{\(\pm 1.3\)}}$ \\
\textbf{SMARTe (Opt Transport)} & 
90.7$_{\scalebox{0.4}{\(\pm 0.4\)}}$ & 
94.5$_{\scalebox{0.4}{\(\pm 0.3\)}}$ & 
94.3$_{\scalebox{0.4}{\(\pm 0.3\)}}$ & 
90.6$_{\scalebox{0.4}{\(\pm 0.4\)}}$ & 
93.6$_{\scalebox{0.4}{\(\pm 0.3\)}}$ & 
94.4$_{\scalebox{0.4}{\(\pm 0.5\)}}$ & 
96.2$_{\scalebox{0.4}{\(\pm 0.4\)}}$ & 
90.7$_{\scalebox{0.4}{\(\pm 0.8\)}}$ \\
\textbf{SMARTe$^{\dagger}$ (Opt Transport)} & 90.7 & \underline{95.1} & \underline{94.7} & 90.7 & \underline{94.2} & \underline{94.3} & \underline{96.5} & 91.4 \\
\bottomrule
\end{tabular}
\label{tab:nyt_overlap}
\end{table*}

\setlength{\tabcolsep}{4pt}
\begin{table*}[ht]
\centering
\caption{F1-score (\%) on sentences with different overlapping patterns and triple numbers for WebNLG* test set. ${\dagger}$ represents the best-performing seed in terms of overall F1 score for reference. $\pm$ represents the standard deviation of the results across 12 runs. *Note: UniRel results are not available for this dataset.}
\begin{tabular}{@{}lllllllll@{}}
\toprule
Model      & Normal & EPO  & SEO  & N=1  & N=2  & N=3  & N=4 & N$\geq$5 \\
\midrule
CasRel \cite{wei2019novel} & 89.4   & 94.7 & 92.2 & 89.3  & 90.8  & 94.2  & 92.4 & 90.9 \\
TPLinker \cite{wang2020tplinker}  & 87.9   & 95.3 & 92.5 & 88.0  & 90.1  & 94.6  & 93.3 & 91.6 \\
PRGC \cite{zheng2021prgc} & 90.4   & 95.9 & 93.6 & 89.9  & 91.6  & 95.0  & \underline{94.8} & 92.8 \\
BiRTE \cite{ren2022simple} & 90.1   & 94.3 & \textbf{95.9} & \underline{90.2}  & \textbf{92.9}  & 95.7  & 94.6 & 92.0 \\
DirectRel \cite{ijcai2022p605} & \textbf{92.0}   & \textbf{97.1} & \underline{94.5} & \textbf{91.6}  & 92.2  & 96.0 & \textbf{95.0}  & \textbf{94.9} \\
SPN \cite{sui2023joint} & - & - & - & 89.5 & 91.3  & \underline{96.4} & 94.7 & \underline{93.8} \\
\hline
SMARTe (Softmax) & 
89.3$_{\scalebox{0.4}{\(\pm 0.5\)}}$ & 
90.2$_{\scalebox{0.4}{\(\pm 1.2\)}}$ & 
93.3$_{\scalebox{0.4}{\(\pm 0.3\)}}$ & 
89.1$_{\scalebox{0.4}{\(\pm 0.5\)}}$ & 
91.3$_{\scalebox{0.4}{\(\pm 0.8\)}}$ & 
95.1$_{\scalebox{0.4}{\(\pm 0.6\)}}$ & 
93.9$_{\scalebox{0.4}{\(\pm 0.5\)}}$ & 
92.9$_{\scalebox{0.4}{\(\pm 0.7\)}}$ \\
\textbf{SMARTe (Opt Transport)} & 
90.5$_{\scalebox{0.4}{\(\pm 0.9\)}}$ & 
90.1$_{\scalebox{0.4}{\(\pm 1.0\)}}$ & 
94.0$_{\scalebox{0.4}{\(\pm 0.3\)}}$ & 
90.1$_{\scalebox{0.4}{\(\pm 0.8\)}}$ & 
91.8$_{\scalebox{0.4}{\(\pm 0.7\)}}$ & 
96.0$_{\scalebox{0.4}{\(\pm 0.5\)}}$ & 
94.8$_{\scalebox{0.4}{\(\pm 0.5\)}}$ & 
93.2$_{\scalebox{0.4}{\(\pm 0.7\)}}$ \\
\textbf{SMARTe$^\dagger$ (Opt Transport)} & \underline{90.6} & 90.6 & \underline{94.5} & 90.0  & \underline{92.4}  & \textbf{96.9}  & \textbf{95.0} & \underline{93.8} \\
\bottomrule
\end{tabular}
\label{tab:webnlg_overlap}
\end{table*}
We present our experimental results for the NYT dataset in Table \ref{tab:nyt_results} and the WebNLG dataset in Table \ref{tab:webnlg_results}, evaluating on both partial and exact matching criteria. Additionally, Table \ref{tab:nyt_overlap} and Table \ref{tab:webnlg_overlap} provide an analysis of our model's performance on overlapping patterns and varying numbers of triples for the NYT and WebNLG datasets, respectively. 

Our results are benchmarked against the latest state-of-the-art (SOTA) methods from 2020 onwards, as we find comparisons with outdated models serve limited purpose beyond formality. To ensure a fair comparison, all models including ours, use \textbf{BERT-Base-Cased} as encoder, with all comparative results sourced directly from their original publications. For our model, we report the \textbf{mean and standard deviation across multiple runs} with our provided seeds in our previous section. Moreover, we also include the \textbf{best-performing seed in terms of overall F1 score for reference (indicated with $\dagger$)}, recognizing that many related works do not explicitly state whether their results are derived from experiments using multiple random seeds or from optimizing the F1 score with a single specific seed. Furthermore, we include results for the softmax variant as a comparison to the optimal transport approach (labeled as ``Opt Transport''). We include zero-shot prompting results from recent LLMs to assess their inherent reasoning capabilities without task-specific fine-tuning, which also reflects a more common and practical usage scenario. Few-shot prompting proves impractical in our context due to extensive relation types (100+ for WebNLG) and multiple relations per text, which complicate prompt design. Fine-tuning LLMs in the same way as our model would offer little advantage over fine-tuning BERT, as it effectively reduces LLMs to encoder backbones and neglects their key strengths such as in-context learning and zero-shot generalization. We therefore include zero-shot LLM results as a \textbf{reference point rather than a direct competitive comparison}, acknowledging the fundamental differences in paradigm and optimization objectives. Note this assessment applies only to Exact matching, as LLMs do not inherently identify the annotated head word required for Partial matching.

Our SMARTe model achieves highly competitive performance, \textbf{coming within 1\% of the top-performing methods and outperforming most previous attempts, while offering the added advantage of interpretability}. We also acknowledge that our model may exhibit a slightly higher dependency on larger datasets compared to SOTA methods as this became evident in the WebNLG exact matching benchmark, where our model's performance was comparatively lower. We attribute this to the presence of numerous poorly defined relationships in the dataset, as illustrated in Figure \ref{fig:webnlg_exact_train}, many of which have fewer than 10 training examples. Additionally, it is also important to note that many previous models were not benchmarked against this variant, making the comparison less conclusive. Furthermore, our analysis shows that the optimal transport algorithm consistently outperforms the softmax approach, particularly in handling overlapping patterns, where the softmax approach shows significantly weaker performance. These findings justify the integration of optimal transport into our model for enhanced performance.
\section{Qualitative Analysis of Explanations}
The qualitative analysis of our SMARTe model demonstrates its capability to accurately predict relational triples while providing interpretable explanations. In Figure \ref{fig:explanation_company}, Slot 7 successfully identifies the relational triple \texttt{(Weinstein, /business/person/company, Films)} by focusing on semantically relevant tokens, such as the subject-object pair and \texttt{founders}, which aligns directly with the relationship \texttt{/business/person/company}. Unlike other slots that generate random attention patterns and fail to predict valid relationships, Slot 7 exhibits a clear and focused attention map. This highlights its ability to isolate critical tokens necessary for prediction, validating the model’s decision-making process while offering valuable insights into its reasoning.

Importantly, this reasoning extends beyond merely predicting \texttt{(Weinstein, founders, Films)}. The model demonstrates a deeper understanding by inferring that \texttt{founders} is semantically tied to the relationship \texttt{/business/person/company}, capturing the relational context effectively. Such interpretability is not an isolated occurrence; similar patterns of focused attention and explainable reasoning have been consistently observed across diverse examples. For instance, entities such as \texttt{\{coach, salary, chef, president, executive\}} are systematically grouped under the \texttt{/business/person/company} relationship, as shown in Figure \ref{fig:snippets_explanation}.

Beyond analyzing successful predictions, we are also interested in examining explanations behind incorrect predictions. For example, in NYT* test sentence 8: \textit{``Mary L. Schapiro, who earlier this year became the new head of NASD, was more amenable to fashioning a deal to the New York Exchange's liking than her predecessor, Robert R. Glauber.''} The ground truth is \texttt{(Glauber, /business/person/company, NASD)}, but our model predicted \texttt{(Schapiro, /business/person/company, NASD)}. This prediction is not entirely unreasonable, as Schapiro is correctly identified as the head of NASD, making the prediction logically valid. The model's reasoning is illustrated in Figure \ref{fig:explanation_wrong}, where Schapiro's association with NASD is highlighted. Hence, this demonstrates that explainability can provide valuable insights into model predictions and help identify potential ambiguities or gaps in the dataset, ultimately supporting efforts to improve data quality and annotation consistency.

For location-related relationships, such as \texttt{/location/location/contains}, the attention mechanism predominantly focuses on the object location, as it appears to encapsulate all the necessary information needed to predict the relational triple, as shown in Figure \ref{fig:explanation_contains279} \& \ref{fig:explanation_contains325}. In these instances, the model’s task shifts from deriving relational meaning from the sentence to recognizing pre-existing factual associations. Since such location-based relationships typically represent fixed world knowledge, the annotations often reflect factual truths rather than information inferred from the sentence context. This underscores a key limitation in providing meaningful explanations for certain types of relationships, as they depend more on external knowledge or the training data than on contextual cues within the text.

For relationship types that are not well-trained due to the limited number of instances, such as \texttt{/people/person/ethnicity} and \texttt{/people/ethnicity/people}, as shown in the training data statistics in Figure \ref{fig:nyt_partial_train}, the model struggles to generate reliable explanations. Although the model correctly predicts the relationship, the generated explanations are incoherent and fail to provide meaningful insights into the model’s reasoning process as seen in Figure \ref{fig:explanation_rare411}.

While explanation quality varies with relationship complexity and characteristics, the model's ability to provide meaningful insights across diverse relationships demonstrates its potential to enhance interpretability and deepen understanding of relational extraction tasks. SMARTe serves as a valuable stepping stone toward truly interpretable relational reasoning in AI systems. Beyond relational triple extraction, this methodology can also be used for a broad range of set-related NLP tasks.

\section{Conclusion}
In this paper, we introduce SMARTe that addresses the critical need for interpretability in relation extraction models. By leveraging slot attention and framing the task as set prediction, SMARTe ensures predictions are explicitly traceable to their learned representations. Our experiments on NYT and WebNLG datasets show that SMARTe achieves performance comparable to state-of-the-art models (within 1\% range) while providing meaningful explanations through slots. These findings demonstrate the feasibility of combining interpretability with effectiveness, addressing the "black-box" limitations of prior approaches. This work responds to the growing necessity for interpretable ML models as NLP systems are increasingly deployed in high-stakes domains requiring accountability, including healthcare \cite{loh2022application}, finance \cite{chen2023explainable}, and law \cite{gorski2021explainable}.

\section{Limitations}
For limitations, we acknowledge that our model's performance slightly lags behind leading models due to its seq2seq design, which does not incorporate combinatorial token interactions like bipartite graphs or interaction tables. Future work will focus on improving token interactions to boost performance while preserving interpretability. Additionally, explanations for complex interactions remain indirect and less intuitive, highlighting differences between model and human text interpretation, which is a known issue with attention-based explanation. To address this, we plan to explore more user-friendly explanation methods to enable users to derive clearer and more actionable insights from predictions.

Another limitation is the computational cost of Hungarian matching, which scales at $O(n^3)$ where n represents our budgeted slot count. While this remains manageable for the current datasets, where sentences typically contain fewer than 10 relational triples and we've allocated sufficient slots, it could become problematic for more complex and longer documents. For scenarios with substantially more relation triples or other NLP objects (i.e., sets of mentions / POS / keywords), Hungarian matching becomes inefficient. Future implementations could explore Chamfer matching as an alternative, which offers reduced $O(n^2)$ complexity while potentially maintaining adequate assignment quality.

\bibliography{acl_latex}

\begin{thebibliography}{37}
\providecommand{\natexlab}[1]{#1}

\bibitem[{Abdin et~al.(2024)Abdin, Aneja, Behl, Bubeck, Eldan, Gunasekar, Harrison, Hewett, Javaheripi, Kauffmann et~al.}]{abdin2024phi}
Marah Abdin, Jyoti Aneja, Harkirat Behl, S{\'e}bastien Bubeck, Ronen Eldan, Suriya Gunasekar, Michael Harrison, Russell~J Hewett, Mojan Javaheripi, Piero Kauffmann, et~al. 2024.
\newblock Phi-4 technical report.
\newblock \emph{arXiv preprint arXiv:2412.08905}.

\bibitem[{Chen et~al.(2023)Chen, Ma, Ren, Lei, Huynh, and Narayan}]{chen2023explainable}
Xun-Qi Chen, Chao-Qun Ma, Yi-Shuai Ren, Yu-Tian Lei, Ngoc Quang~Anh Huynh, and Seema Narayan. 2023.
\newblock Explainable artificial intelligence in finance: A bibliometric review.
\newblock \emph{Finance Research Letters}, page 104145.

\bibitem[{Chen et~al.(2021)Chen, Zhang, Hu, and Huang}]{chen2021jointly}
Yubo Chen, Yunqi Zhang, Changran Hu, and Yongfeng Huang. 2021.
\newblock Jointly extracting explicit and implicit relational triples with reasoning pattern enhanced binary pointer network.
\newblock In \emph{Proceedings of the 2021 Conference of the North American Chapter of the Association for Computational Linguistics: Human Language Technologies}, pages 5694--5703.

\bibitem[{Chung et~al.(2014)Chung, Gulcehre, Cho, and Bengio}]{chung2014empirical}
Junyoung Chung, Caglar Gulcehre, KyungHyun Cho, and Yoshua Bengio. 2014.
\newblock Empirical evaluation of gated recurrent neural networks on sequence modeling.
\newblock \emph{arXiv preprint arXiv:1412.3555}.

\bibitem[{Danilevsky et~al.(2020)Danilevsky, Qian, Aharonov, Katsis, Kawas, and Sen}]{danilevsky2020survey}
Marina Danilevsky, Kun Qian, Ranit Aharonov, Yannis Katsis, Ban Kawas, and Prithviraj Sen. 2020.
\newblock A survey of the state of explainable ai for natural language processing.
\newblock In \emph{Proceedings of the 1st Conference of the Asia-Pacific Chapter of the Association for Computational Linguistics and the 10th International Joint Conference on Natural Language Processing}, pages 447--459.

\bibitem[{Devlin et~al.(2019)Devlin, Chang, Lee, and Toutanova}]{devlin-etal-2019-bert}
Jacob Devlin, Ming-Wei Chang, Kenton Lee, and Kristina Toutanova. 2019.
\newblock \href {https://doi.org/10.18653/v1/N19-1423} {{BERT}: Pre-training of deep bidirectional transformers for language understanding}.
\newblock In \emph{Proceedings of the 2019 Conference of the North {A}merican Chapter of the Association for Computational Linguistics: Human Language Technologies, Volume 1 (Long and Short Papers)}, pages 4171--4186, Minneapolis, Minnesota. Association for Computational Linguistics.

\bibitem[{Fu et~al.(2019)Fu, Li, and Ma}]{fu2019graphrel}
Tsu-Jui Fu, Peng-Hsuan Li, and Wei-Yun Ma. 2019.
\newblock Graphrel: Modeling text as relational graphs for joint entity and relation extraction.
\newblock In \emph{Proceedings of the 57th annual meeting of the association for computational linguistics}, pages 1409--1418.

\bibitem[{Gardent et~al.(2017)Gardent, Shimorina, Narayan, and Perez-Beltrachini}]{gardent2017creating}
Claire Gardent, Anastasia Shimorina, Shashi Narayan, and Laura Perez-Beltrachini. 2017.
\newblock Creating training corpora for nlg micro-planning.
\newblock In \emph{55th Annual Meeting of the Association for Computational Linguistics, ACL 2017}, pages 179--188. Association for Computational Linguistics (ACL).

\bibitem[{Gemma-Team(2025)}]{gemma_2025}
Gemma-Team. 2025.
\newblock \href {https://goo.gle/Gemma3Report} {Gemma 3}.

\bibitem[{G{\'o}rski and Ramakrishna(2021)}]{gorski2021explainable}
{\L}ukasz G{\'o}rski and Shashishekar Ramakrishna. 2021.
\newblock Explainable artificial intelligence, lawyer's perspective.
\newblock In \emph{Proceedings of the eighteenth international conference on artificial intelligence and law}, pages 60--68.

\bibitem[{Grattafiori et~al.(2024)Grattafiori, Dubey, Jauhri, Pandey, Kadian, Al-Dahle, Letman, Mathur, Schelten, Vaughan et~al.}]{grattafiori2024llama}
Aaron Grattafiori, Abhimanyu Dubey, Abhinav Jauhri, Abhinav Pandey, Abhishek Kadian, Ahmad Al-Dahle, Aiesha Letman, Akhil Mathur, Alan Schelten, Alex Vaughan, et~al. 2024.
\newblock The llama 3 herd of models.
\newblock \emph{arXiv preprint arXiv:2407.21783}.

\bibitem[{Gupta et~al.(2016)Gupta, Sch{\"u}tze, and Andrassy}]{gupta2016table}
Pankaj Gupta, Hinrich Sch{\"u}tze, and Bernt Andrassy. 2016.
\newblock Table filling multi-task recurrent neural network for joint entity and relation extraction.
\newblock In \emph{Proceedings of COLING 2016, the 26th International Conference on Computational Linguistics: Technical Papers}, pages 2537--2547.

\bibitem[{Kuhn(1955)}]{kuhn1955hungarian}
Harold~W Kuhn. 1955.
\newblock The hungarian method for the assignment problem.
\newblock \emph{Naval research logistics quarterly}, 2(1-2):83--97.

\bibitem[{Li et~al.(2021)Li, Luo, Dong, Yang, Luan, and He}]{li2021tdeer}
Xianming Li, Xiaotian Luo, Chenghao Dong, Daichuan Yang, Beidi Luan, and Zhen He. 2021.
\newblock Tdeer: An efficient translating decoding schema for joint extraction of entities and relations.
\newblock In \emph{Proceedings of the 2021 conference on empirical methods in natural language processing}, pages 8055--8064.

\bibitem[{Locatello et~al.(2020)Locatello, Weissenborn, Unterthiner, Mahendran, Heigold, Uszkoreit, Dosovitskiy, and Kipf}]{locatello2020object}
Francesco Locatello, Dirk Weissenborn, Thomas Unterthiner, Aravindh Mahendran, Georg Heigold, Jakob Uszkoreit, Alexey Dosovitskiy, and Thomas Kipf. 2020.
\newblock Object-centric learning with slot attention.
\newblock \emph{Advances in neural information processing systems}, 33:11525--11538.

\bibitem[{Loh et~al.(2022)Loh, Ooi, Seoni, Barua, Molinari, and Acharya}]{loh2022application}
Hui~Wen Loh, Chui~Ping Ooi, Silvia Seoni, Prabal~Datta Barua, Filippo Molinari, and U~Rajendra Acharya. 2022.
\newblock Application of explainable artificial intelligence for healthcare: A systematic review of the last decade (2011--2022).
\newblock \emph{Computer Methods and Programs in Biomedicine}, 226:107161.

\bibitem[{Miwa and Sasaki(2014)}]{miwa2014modeling}
Makoto Miwa and Yutaka Sasaki. 2014.
\newblock Modeling joint entity and relation extraction with table representation.
\newblock In \emph{Proceedings of the 2014 conference on empirical methods in natural language processing (EMNLP)}, pages 1858--1869.

\bibitem[{Nayak et~al.(2021)Nayak, Majumder, Goyal, and Poria}]{nayak2021deep}
Tapas Nayak, Navonil Majumder, Pawan Goyal, and Soujanya Poria. 2021.
\newblock Deep neural approaches to relation triplets extraction: A comprehensive survey.
\newblock \emph{Cognitive Computation}, 13(5):1215--1232.

\bibitem[{OpenAI et~al.(2024)OpenAI, Achiam, Adler, and Others}]{openai2024gpt4technicalreport}
OpenAI, Josh Achiam, Steven Adler, and Others. 2024.
\newblock \href {https://arxiv.org/abs/2303.08774} {Gpt-4 technical report}.
\newblock \emph{Preprint}, arXiv:2303.08774.

\bibitem[{Qwen-Team(2025)}]{qwen3}
Qwen-Team. 2025.
\newblock \href {https://qwenlm.github.io/blog/qwen3/} {Qwen3}.

\bibitem[{Ren et~al.(2022)Ren, Zhang, Zhao, Yin, Liu, and Li}]{ren2022simple}
Feiliang Ren, Longhui Zhang, Xiaofeng Zhao, Shujuan Yin, Shilei Liu, and Bochao Li. 2022.
\newblock A simple but effective bidirectional framework for relational triple extraction.
\newblock In \emph{Proceedings of the fifteenth ACM international conference on web search and data mining}, pages 824--832.

\bibitem[{Riedel et~al.(2010)Riedel, Yao, and McCallum}]{riedel2010modeling}
Sebastian Riedel, Limin Yao, and Andrew McCallum. 2010.
\newblock Modeling relations and their mentions without labeled text.
\newblock In \emph{Machine Learning and Knowledge Discovery in Databases: European Conference, ECML PKDD 2010, Barcelona, Spain, September 20-24, 2010, Proceedings, Part III 21}, pages 148--163. Springer.

\bibitem[{Roth and Yih(2004)}]{roth2004linear}
Dan Roth and Wen-tau Yih. 2004.
\newblock A linear programming formulation for global inference in natural language tasks.
\newblock In \emph{Proceedings of the eighth conference on computational natural language learning (CoNLL-2004) at HLT-NAACL 2004}, pages 1--8.

\bibitem[{Shang et~al.(2022)Shang, Huang, Sun, Wei, and Mao}]{ijcai2022p605}
Yu-Ming Shang, Heyan Huang, Xin Sun, Wei Wei, and Xian-Ling Mao. 2022.
\newblock \href {https://doi.org/10.24963/ijcai.2022/605} {Relational triple extraction: One step is enough}.
\newblock In \emph{Proceedings of the Thirty-First International Joint Conference on Artificial Intelligence, {IJCAI-22}}, pages 4360--4366. International Joint Conferences on Artificial Intelligence Organization.
\newblock Main Track.

\bibitem[{Sui et~al.(2023)Sui, Zeng, Chen, Liu, and Zhao}]{sui2023joint}
Dianbo Sui, Xiangrong Zeng, Yubo Chen, Kang Liu, and Jun Zhao. 2023.
\newblock Joint entity and relation extraction with set prediction networks.
\newblock \emph{IEEE Transactions on Neural Networks and Learning Systems}.

\bibitem[{Tang et~al.(2022)Tang, Xu, Zhao, Mao, Liu, Liao, and Xie}]{tang2022unirel}
Wei Tang, Benfeng Xu, Yuyue Zhao, Zhendong Mao, Yifeng Liu, Yong Liao, and Haiyong Xie. 2022.
\newblock Unirel: Unified representation and interaction for joint relational triple extraction.
\newblock In \emph{Proceedings of the 2022 Conference on Empirical Methods in Natural Language Processing}, pages 7087--7099.

\bibitem[{Villani et~al.(2009)}]{villani2009optimal}
C{\'e}dric Villani et~al. 2009.
\newblock \emph{Optimal transport: old and new}, volume 338.
\newblock Springer.

\bibitem[{Wang et~al.(2020)Wang, Yu, Zhang, Liu, Zhu, and Sun}]{wang2020tplinker}
Yucheng Wang, Bowen Yu, Yueyang Zhang, Tingwen Liu, Hongsong Zhu, and Limin Sun. 2020.
\newblock Tplinker: Single-stage joint extraction of entities and relations through token pair linking.
\newblock In \emph{Proceedings of the 28th International Conference on Computational Linguistics}, pages 1572--1582.

\bibitem[{Wei et~al.(2019)Wei, Su, Wang, Tian, and Chang}]{wei2019novel}
Zhepei Wei, Jianlin Su, Yue Wang, Yuan Tian, and Yi~Chang. 2019.
\newblock A novel cascade binary tagging framework for relational triple extraction.
\newblock \emph{arXiv preprint arXiv:1909.03227}.

\bibitem[{Yan et~al.(2021)Yan, Zhang, Fu, Zhang, and Wei}]{yan2021partition}
Zhiheng Yan, Chong Zhang, Jinlan Fu, Qi~Zhang, and Zhongyu Wei. 2021.
\newblock A partition filter network for joint entity and relation extraction.
\newblock In \emph{Proceedings of the 2021 Conference on Empirical Methods in Natural Language Processing}, pages 185--197.

\bibitem[{Ye et~al.(2021)Ye, Zhang, Deng, Chen, Tan, Huang, and Chen}]{ye2021contrastive}
Hongbin Ye, Ningyu Zhang, Shumin Deng, Mosha Chen, Chuanqi Tan, Fei Huang, and Huajun Chen. 2021.
\newblock Contrastive triple extraction with generative transformer.
\newblock In \emph{Proceedings of the AAAI conference on artificial intelligence}, volume~35, pages 14257--14265.

\bibitem[{Zeng et~al.(2020)Zeng, Zhang, and Liu}]{zeng2020copymtl}
Daojian Zeng, Haoran Zhang, and Qianying Liu. 2020.
\newblock Copymtl: Copy mechanism for joint extraction of entities and relations with multi-task learning.
\newblock In \emph{Proceedings of the AAAI conference on artificial intelligence}, volume~34, pages 9507--9514.

\bibitem[{Zeng et~al.(2018)Zeng, Zeng, He, Liu, and Zhao}]{zeng2018extracting}
Xiangrong Zeng, Daojian Zeng, Shizhu He, Kang Liu, and Jun Zhao. 2018.
\newblock Extracting relational facts by an end-to-end neural model with copy mechanism.
\newblock In \emph{Proceedings of the 56th Annual Meeting of the Association for Computational Linguistics (Volume 1: Long Papers)}, pages 506--514.

\bibitem[{Zhang et~al.(2017)Zhang, Zhang, and Fu}]{zhang2017end}
Meishan Zhang, Yue Zhang, and Guohong Fu. 2017.
\newblock End-to-end neural relation extraction with global optimization.
\newblock In \emph{Proceedings of the 2017 conference on empirical methods in natural language processing}, pages 1730--1740.

\bibitem[{Zhang et~al.(2023)Zhang, Zhang, Lacoste-Julien, Burghouts, and Snoek}]{pmlr-v202-zhang23ba}
Yan Zhang, David~W. Zhang, Simon Lacoste-Julien, Gertjan~J. Burghouts, and Cees G.~M. Snoek. 2023.
\newblock \href {https://proceedings.mlr.press/v202/zhang23ba.html} {Unlocking slot attention by changing optimal transport costs}.
\newblock In \emph{Proceedings of the 40th International Conference on Machine Learning}, volume 202 of \emph{Proceedings of Machine Learning Research}, pages 41931--41951. PMLR.

\bibitem[{Zheng et~al.(2021)Zheng, Wen, Chen, Yang, Zhang, Zhang, Zhang, Qin, Xu, and Zheng}]{zheng2021prgc}
Hengyi Zheng, Rui Wen, Xi~Chen, Yifan Yang, Yunyan Zhang, Ziheng Zhang, Ningyu Zhang, Bin Qin, Ming Xu, and Yefeng Zheng. 2021.
\newblock Prgc: Potential relation and global correspondence based joint relational triple extraction.
\newblock \emph{arXiv preprint arXiv:2106.09895}.

\bibitem[{Zheng et~al.(2017)Zheng, Wang, Bao, Hao, Zhou, and Xu}]{zheng2017joint}
Suncong Zheng, Feng Wang, Hongyun Bao, Yuexing Hao, Peng Zhou, and Bo~Xu. 2017.
\newblock Joint extraction of entities and relations based on a novel tagging scheme.
\newblock In \emph{Proceedings of the 55th Annual Meeting of the Association for Computational Linguistics (Volume 1: Long Papers)}, pages 1227--1236.

\end{thebibliography}
\clearpage
\appendix
\onecolumn
\section{Analysis of our SMARTe model Explanations}
\label{app:A}
\begin{figure*}[hbt!]
\centering
  \includegraphics[width=0.8\linewidth]{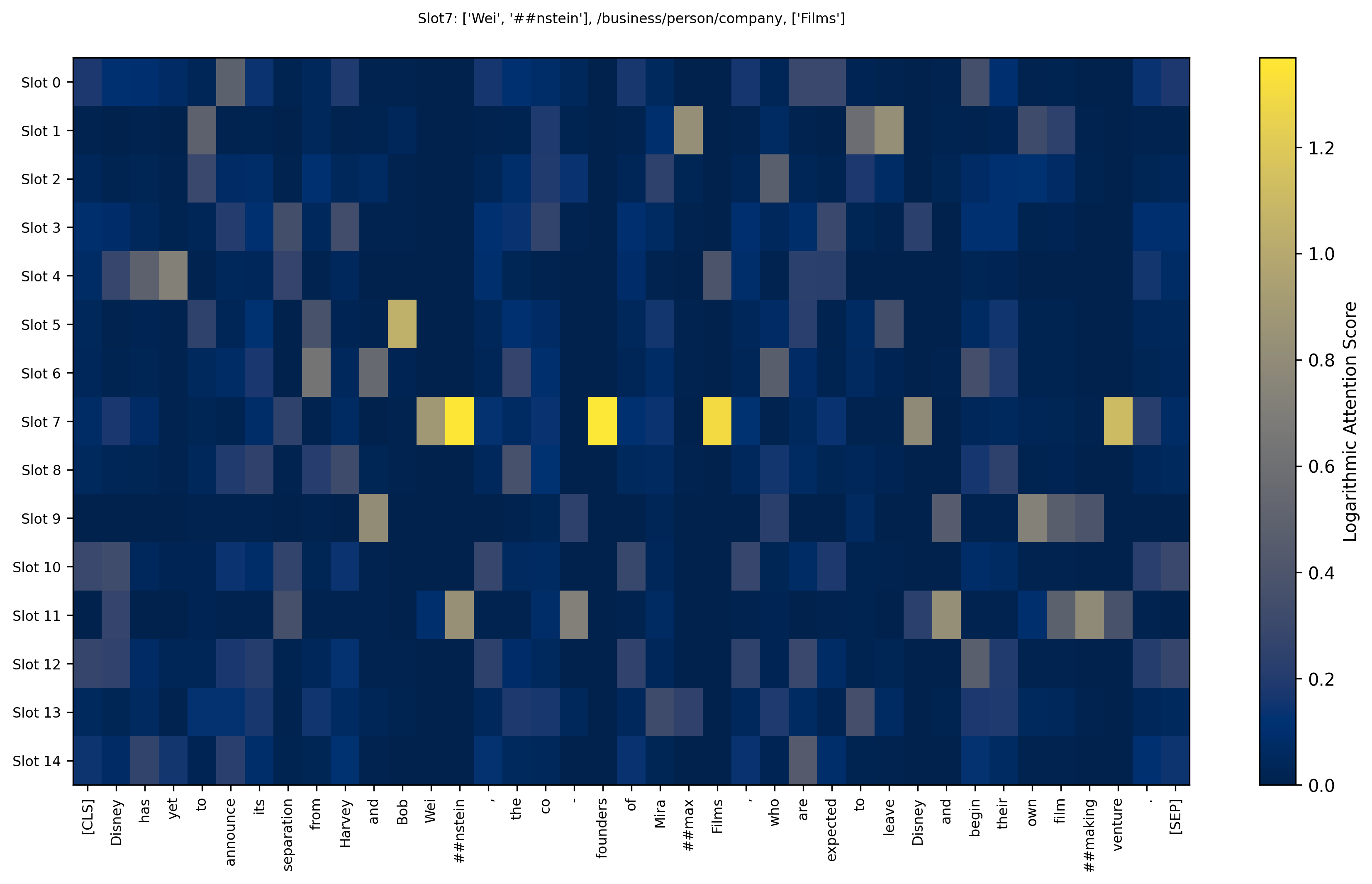}
  \caption{Visualization of the logarithmic attention scores for each token across all slots for the prediction in NYT partial matching test set sentence 429. Slot 7 successfully predicts the relational triple (\texttt{Weinstein}, \texttt{/business/person/company}, \texttt{Films}), while other slots yield no valid predictions (classified as NAs). The contributing tokens are highlighted, with the model assigning high attention scores to tokens such as its subject / object pair and \texttt{founders} which aligns with the relationship \texttt{/business/person/company}.} 
  \label{fig:explanation_company}
\end{figure*}
\begin{figure*}[hbt!]
\centering
  \includegraphics[width=0.8\linewidth]{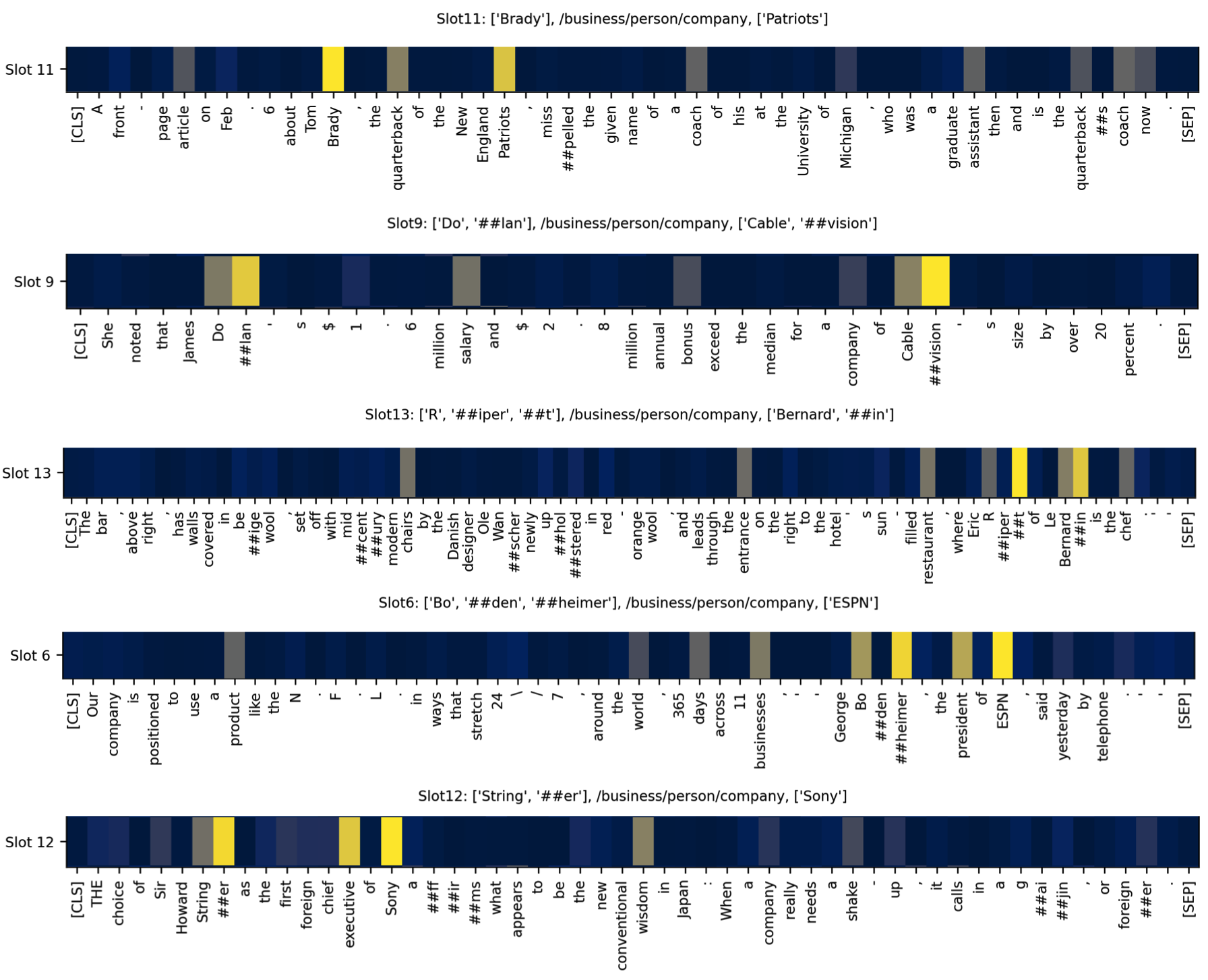}
  \caption{Snippets of explanation in NYT partial matching test set for various sentences: 188,196, 263, 349, 371 for \texttt{/business/person/company} relational triple.} 
  \label{fig:snippets_explanation}
\end{figure*}

\begin{figure*}[hbt!]
\centering
  \includegraphics[width=0.9\linewidth]{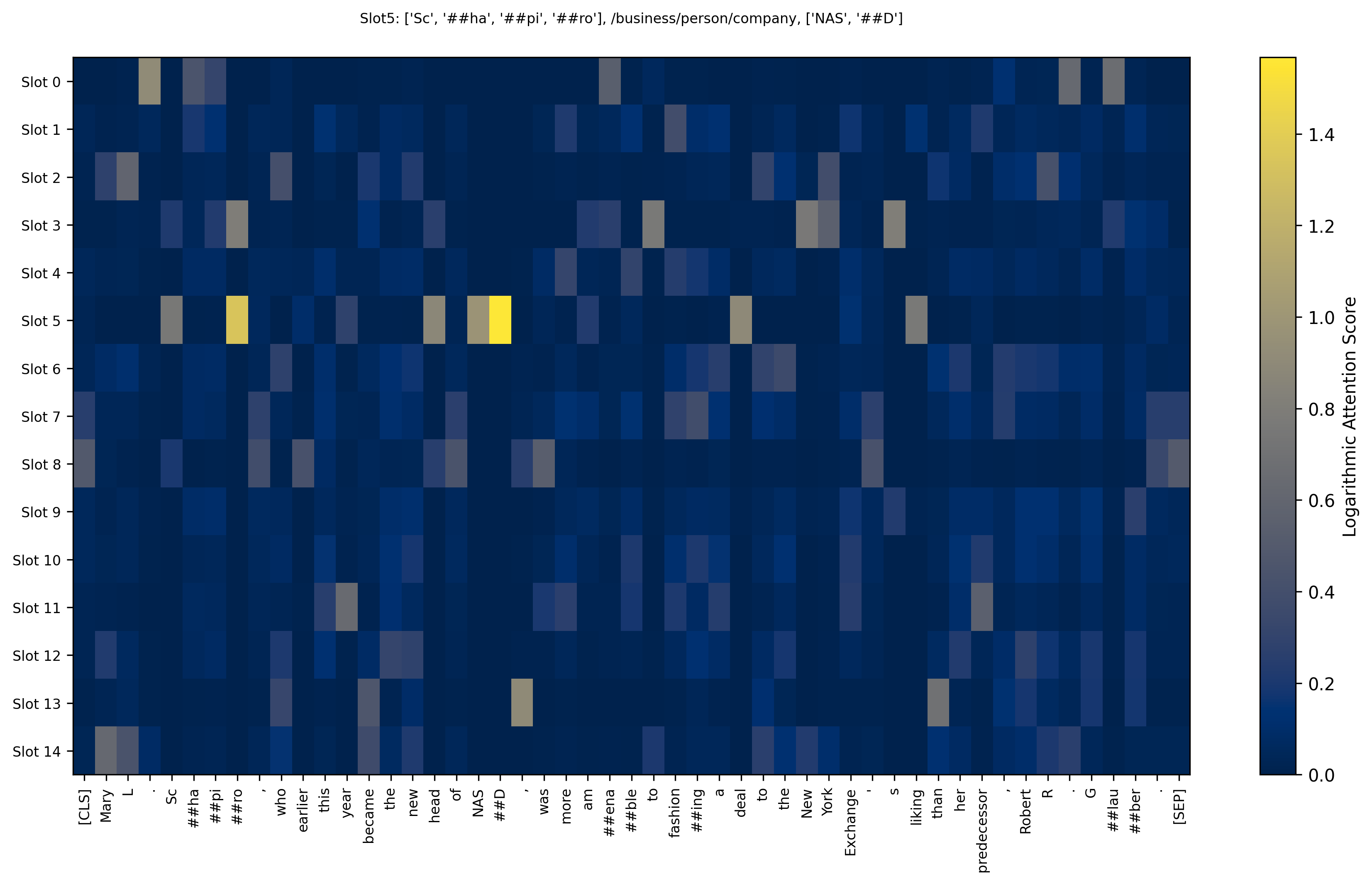}
  \caption{Visualization of incorrectly predicted relational triple for NYT partial matching test sentence 8. The golden triple for the sentence is \texttt{(Glauber, /business/person/company, NASD)}. However, our model predicted \texttt{(Schapiro, /business/person/company, NASD)}. This misprediction occurred because the model identified Schapiro as being associated with NASD in a valid context, as Schapiro held a leadership (\texttt{head}) position in the organization. Although this reasoning is correct, the prediction does not match the ground truth, leading to an incorrect result under the evaluation criteria.} 
  \label{fig:explanation_wrong}
\end{figure*}

\begin{figure*}[hbt!]
\centering
  \includegraphics[width=0.9\linewidth]{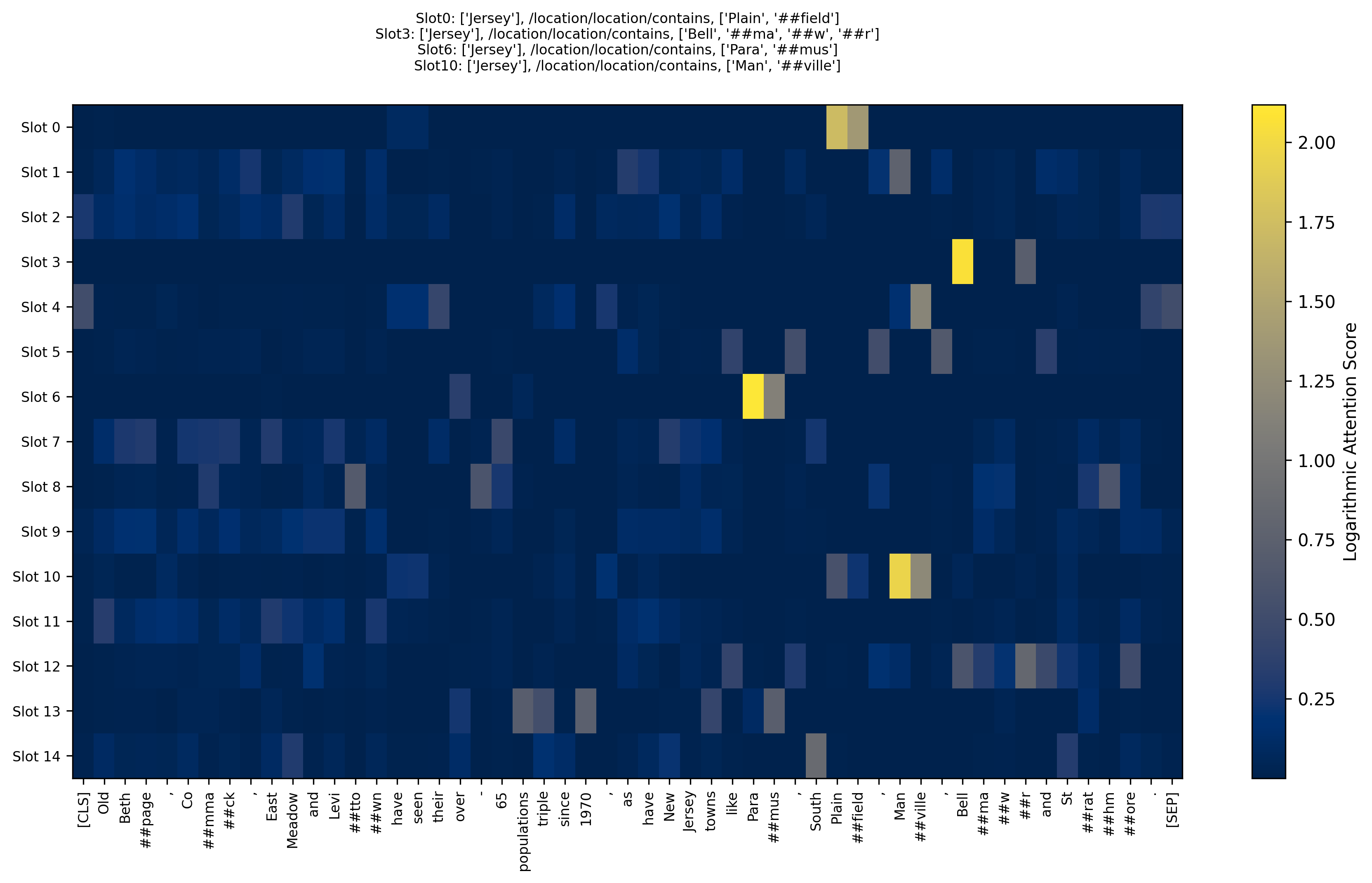}
  \caption{Explanations for NYT partial matching test sentence 279 show that the attentions primarily highlight the object location for the relationship \texttt{/location/location/contains}, where information from the object location alone is sufficient to predict the relational triple. In this sentence, Jersey contains Plainfield, Bellmawr, Paramus, and Manville.} 
  \label{fig:explanation_contains279}
\end{figure*}

\begin{figure*}[hbt!]
\centering
  \includegraphics[width=0.9\linewidth]{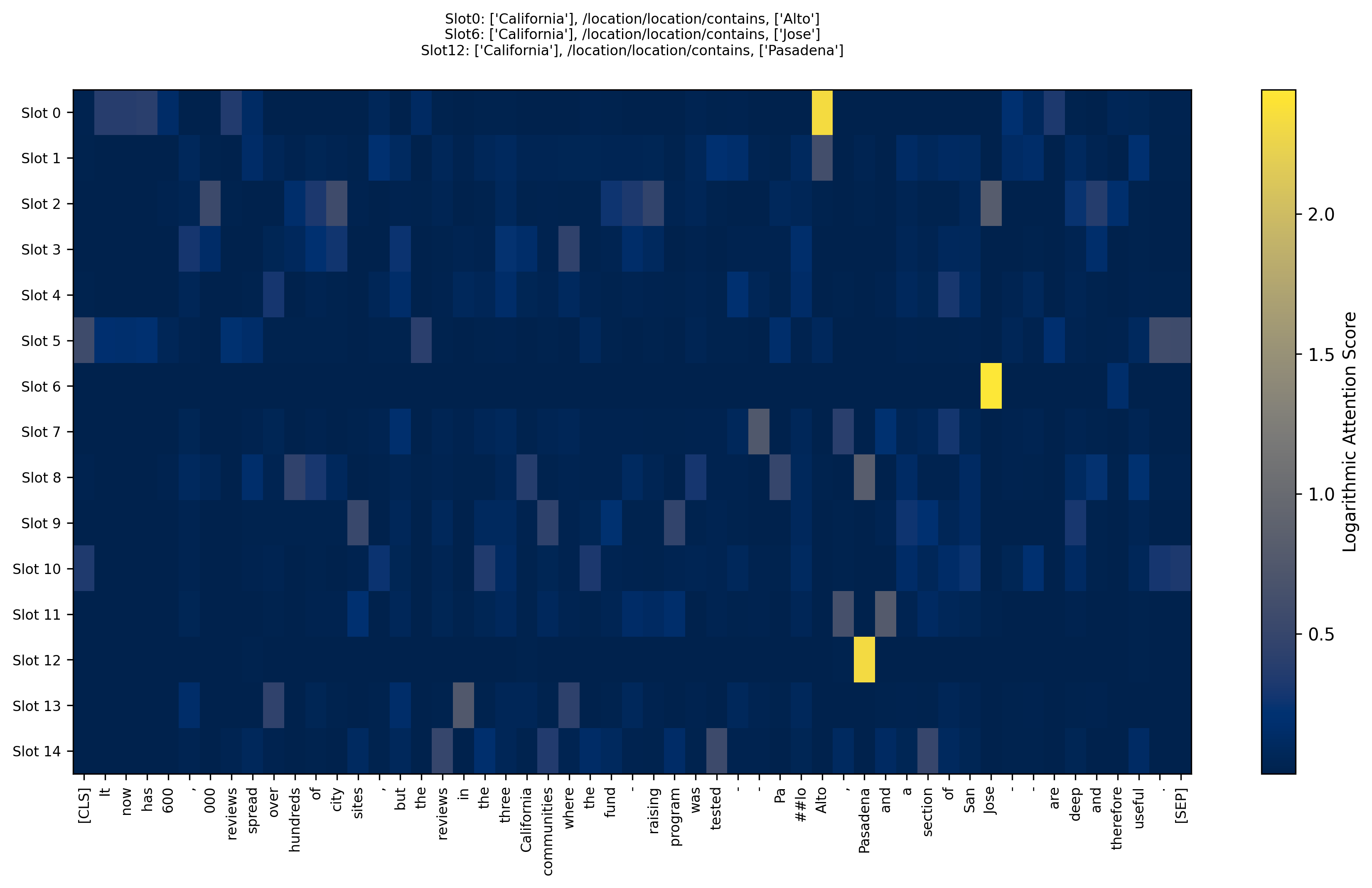}
  \caption{Explanations for NYT partial matching test sentence 325 show that the attentions primarily highlight the object location for the relationship \texttt{/location/location/contains}, where information from the object location alone is sufficient to predict the relational triple. In this sentence, California contains Alto, Jose, Pasadena.} 
  \label{fig:explanation_contains325}
\end{figure*}

\begin{figure*}[hbt!]
\centering
  \includegraphics[width=0.9\linewidth]{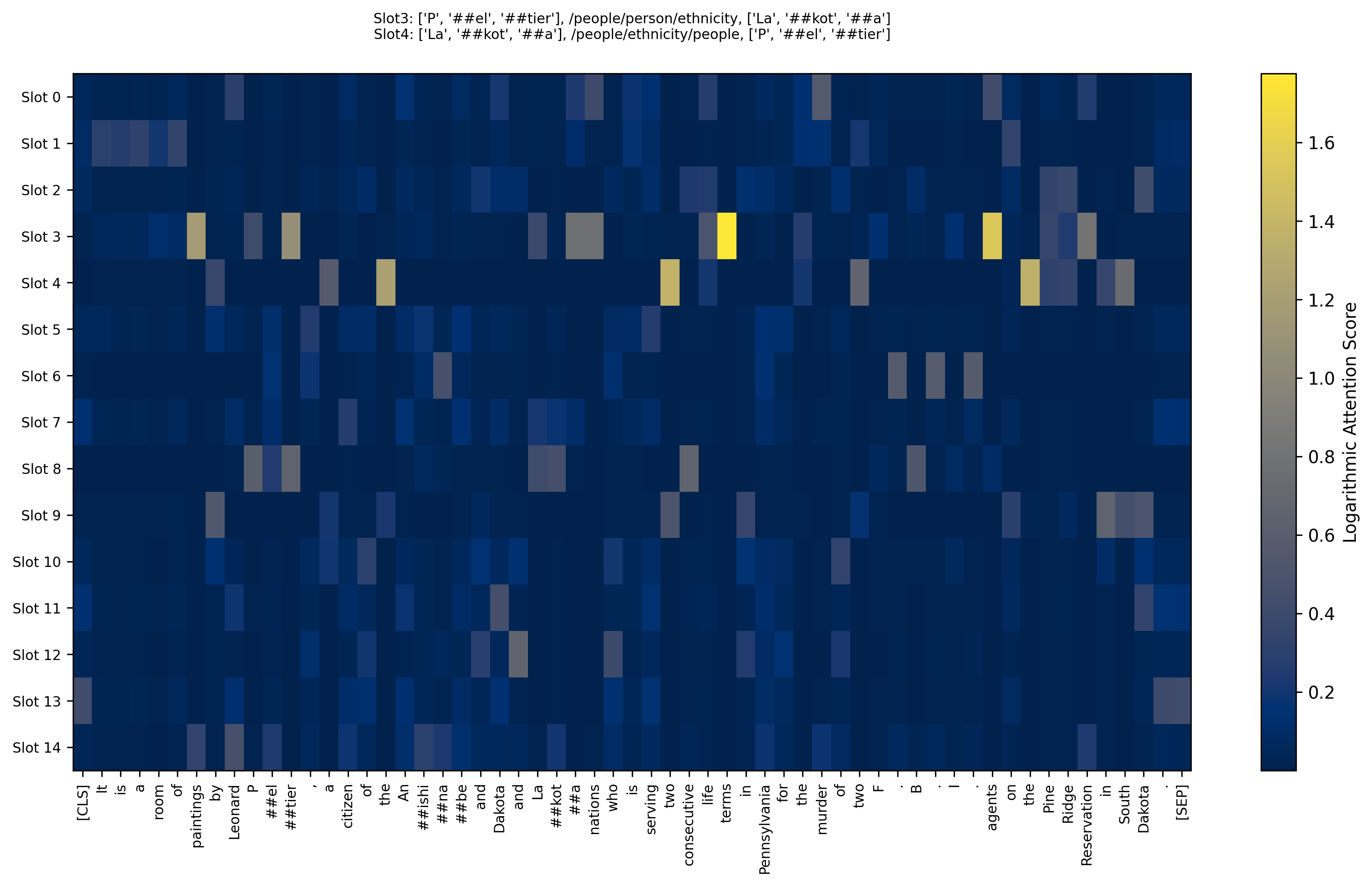}
  \caption{Explanations for NYT partial matching test sentence 411: While the model correctly predicts the relationship, the generated explanations are incoherent and lack meaningful insights into the model’s reasoning process, likely due to the scarcity of training instances for these specific relationships.} 
  \label{fig:explanation_rare411}
\end{figure*}

\clearpage

\section{Statistics of NYT and WebNLG datasets}
\begin{table*}[hbt!]
\centering
\caption{Statistics of the datasets used in the study. The symbol (*) denotes datasets involving partial matching, whereas datasets without this symbol correspond to exact matching. $^{\ddagger}$ relations not in train dataset are removed.}
\label{tab:dataset_statistics}
\begin{tabular}{lccccccccc}
\toprule
\multirow{3}{*}{Category} & \multicolumn{7}{c}{Dataset} \\
\cmidrule(lr){2-8}
& \multicolumn{2}{c}{Train} & \multicolumn{2}{c}{Valid} & \multicolumn{2}{c}{Test} & \multirow{2}{*}{Relations}\\
\cmidrule(lr){2-3} \cmidrule(lr){4-5} \cmidrule(lr){6-7}
& Sents & Triples & Sents & Triples & Sents & Triples & \\
\midrule
NYT* & 56195 & 88253 & 4999 & 7976 & 5000 & 8110 & 24 \\
NYT & 56196 & 88366 & 5000 & 7985 & 5000 & 8120 & 24 \\
WebNLG* & 5019 & 11776 & 500 & 1117 & 703 & 1591 & 170$^{\ddagger}$ \\
WebNLG & 5019 & 11313 & 500 & 1224 & 703 & 1607 & 211$^{\ddagger}$ \\
\bottomrule
\end{tabular}
\end{table*}
\begin{table*}[hbt!]
\centering
\caption{Statistics of the overlapping patterns \cite{zeng2018extracting} across Train, Valid, and Test sets, following prior work. Please note that this form of evaluation is only applicable to partial matching datasets.}
\label{tab:overlapping_statistics}
\begin{tabular}{lccccccccccc}
\toprule
\multirow{2}{*}{Category} & \multicolumn{3}{c}{Train Set} & \multicolumn{3}{c}{Valid Set} & \multicolumn{3}{c}{Test Set} \\
\cmidrule(lr){2-4} \cmidrule(lr){5-7} \cmidrule(lr){8-10}
& Normal & SEO & EPO & Normal & SEO & EPO & Normal & SEO & EPO \\
\midrule
NYT* & 37013 & 14735 & 9782 & 3306 & 1350 & 849 & 3266 & 1297 & 978\\
WebNLG* & 1600 & 3402 & 227 & 182 & 318 & 16 & 246 & 457 & 26\\
\bottomrule
\end{tabular}
\end{table*}
\clearpage
\begin{figure*}[hbt!]
\centering
  \includegraphics[width=0.8\linewidth]{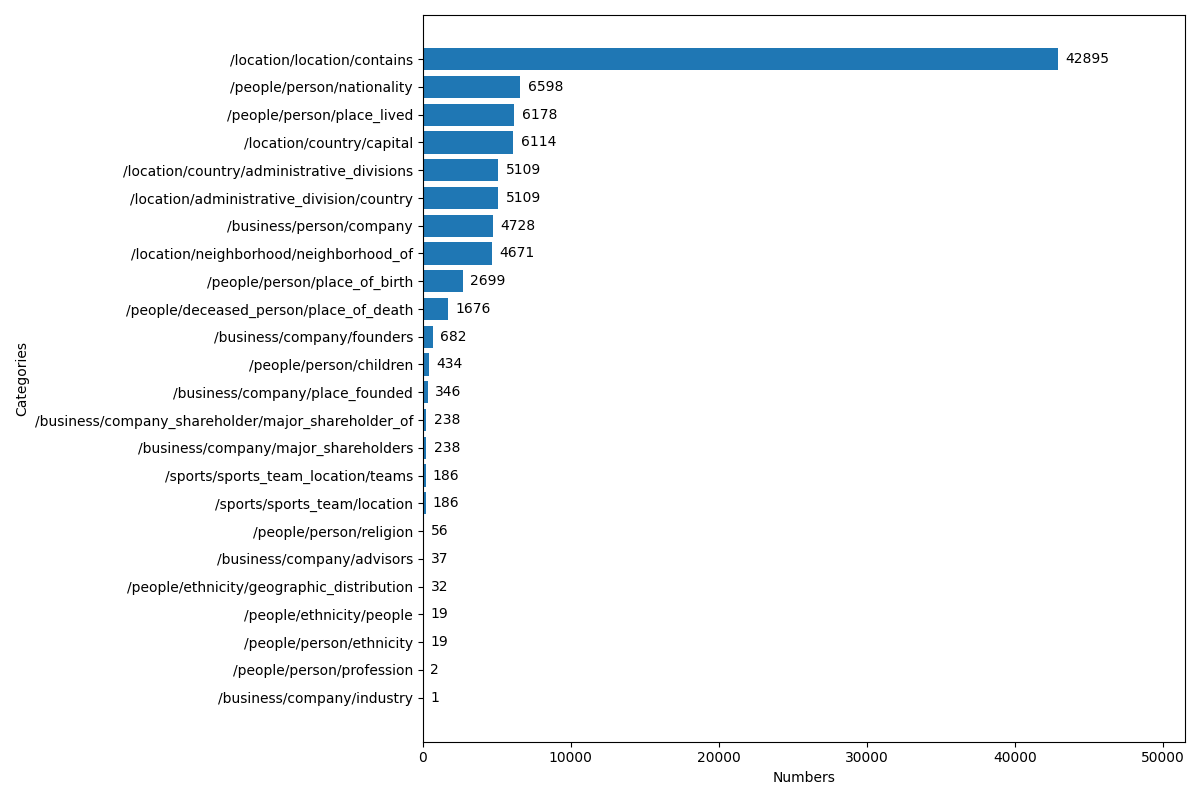}
  \caption{NYT partial matching train dataset relationship statistics} 
  \label{fig:nyt_partial_train}
\end{figure*}
\begin{figure*}[hbt!]
\centering
  \includegraphics[width=0.8\linewidth]{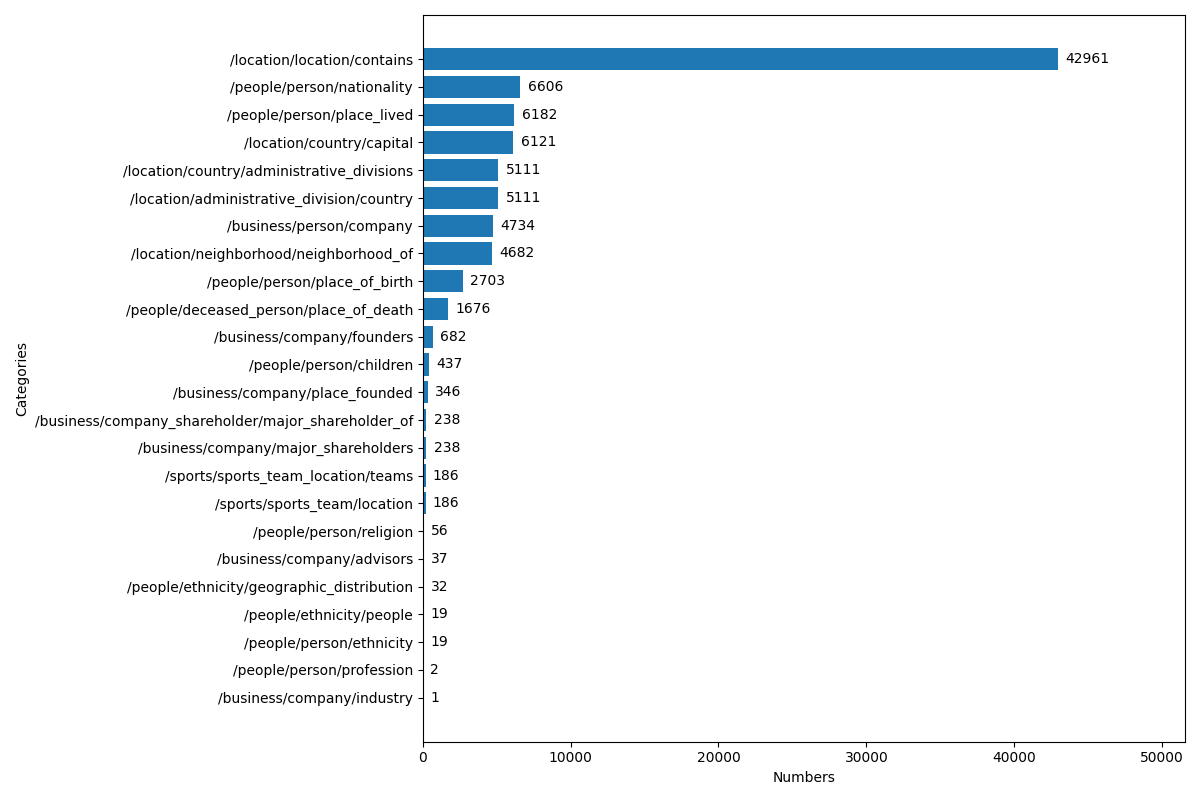}
  \caption{NYT exact matching train dataset relationship statistics} 
  \label{fig:nyt_exact_train}
\end{figure*}
\begin{figure*}[hbt!]
\centering
  \includegraphics[width=0.8\linewidth]{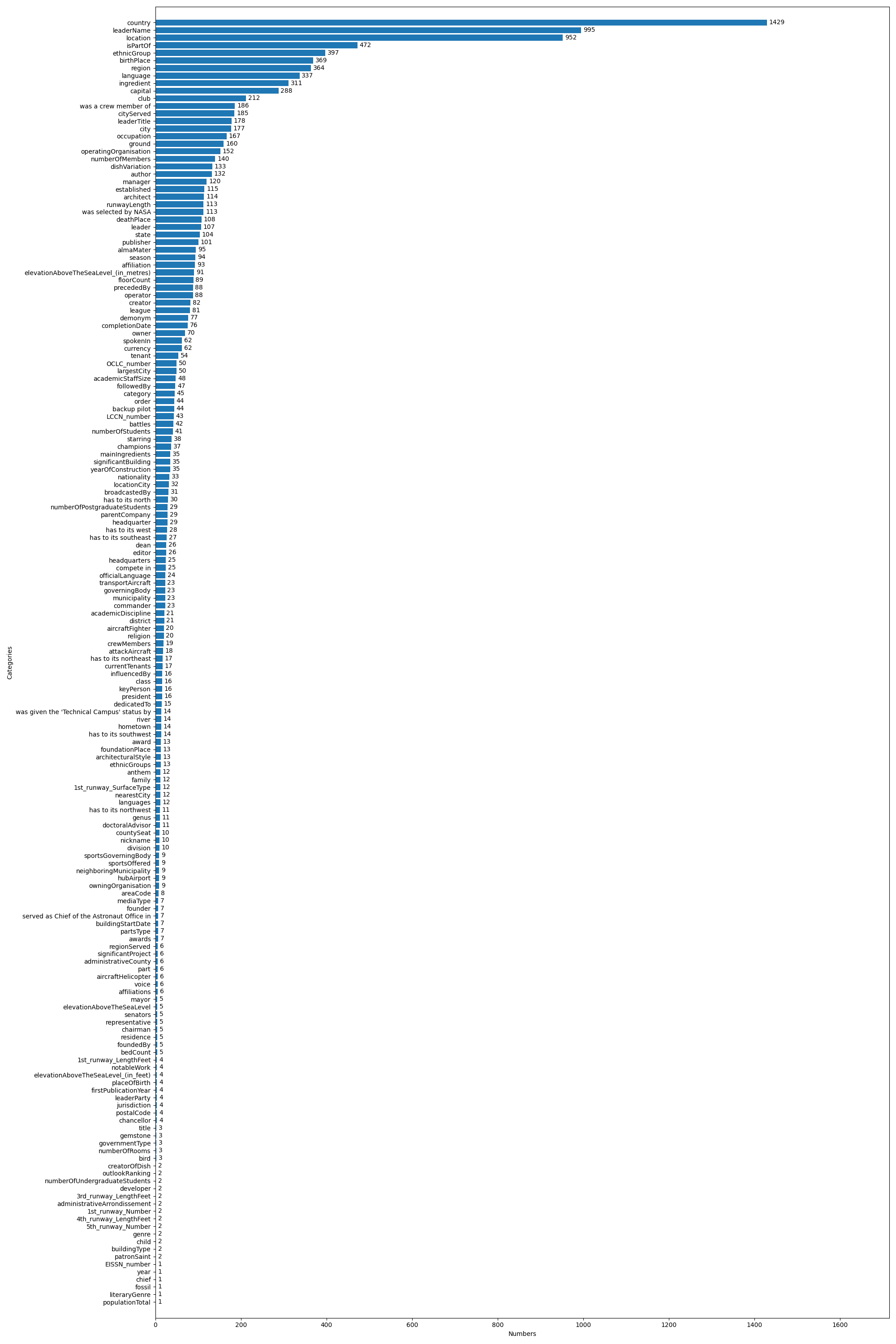}
  \caption{WebNLG partial matching train dataset relationship statistics} 
  \label{fig:webnlg_partial_train}
\end{figure*}
\begin{figure*}[hbt!]
\centering
  \includegraphics[width=0.8\linewidth]{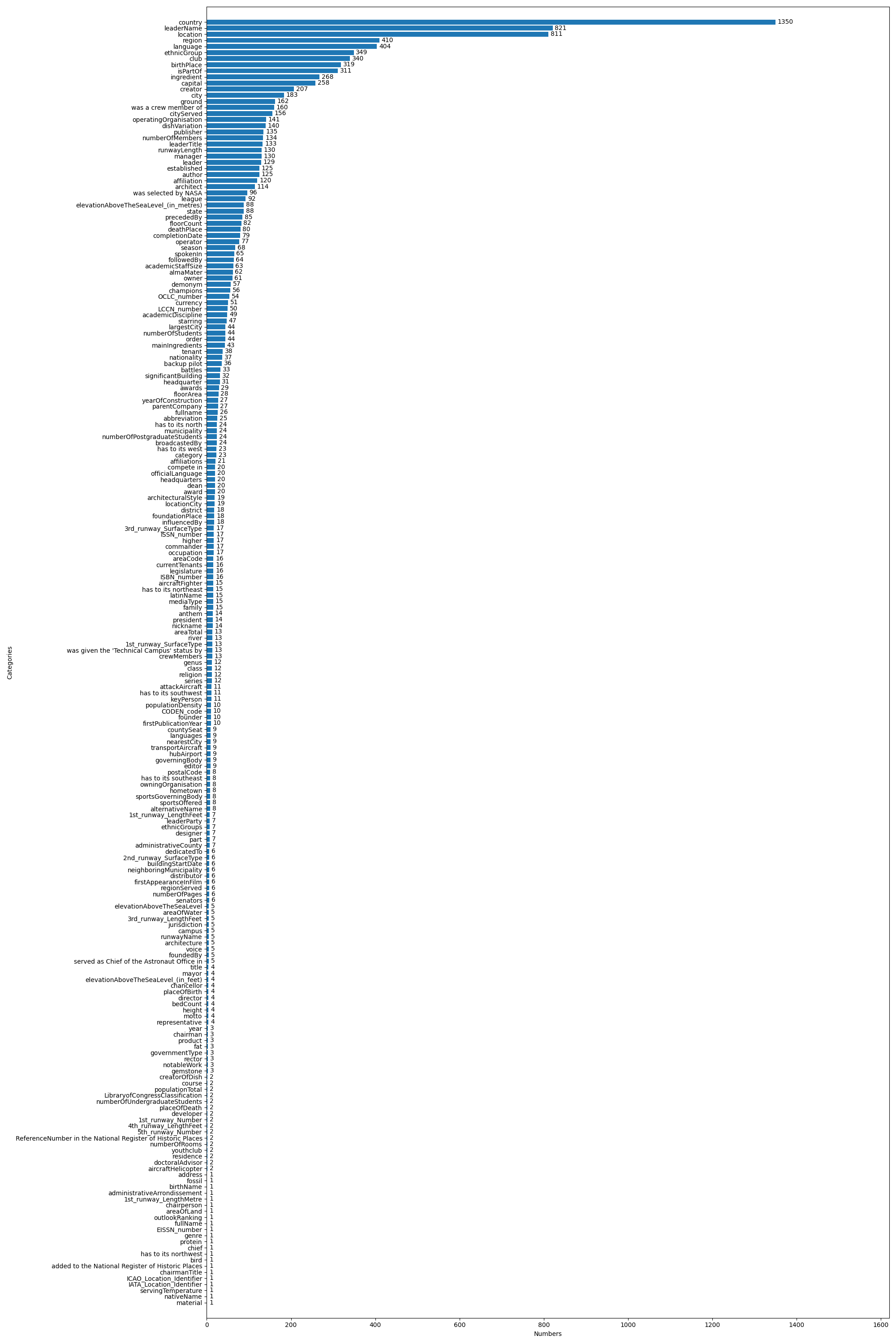}
  \caption{WebNLG exact matching train dataset relationship statistics} 
  \label{fig:webnlg_exact_train}
\end{figure*}
\clearpage

\section{Experiment Hyperparameter Configuration}

Our experiments were conducted on a GeForce RTX 2080 Ti GPU (11GB VRAM) and ran with 12 random seeds (11, 29, 33, 39, 42, 53, 57, 62, 65, 73, 96, 98). Training time per epoch is approximately 7 minutes for NYT and 1 minute for WebNLG.
 
\label{app:C}
\begin{table*}[hbt!]
    \centering
    \begin{tabular}{lcccc}
        \toprule
        Parameter & NYT* & NYT & WebNLG* & WebNLG \\
        \midrule
        batch\_size & 8 & 8 & 8 & 8 \\
        epochs & 80 & 100 & 340 & 340 \\
        num\_classes & 25 & 25 & 170 & 211 \\
        num\_generated\_triples & 15 & 15 & 15 & 15 \\
        optimizer & AdamW & AdamW & AdamW & AdamW \\
        encoder\_lr & 2.00E-05 & 1.00E-05 & 2.00E-05 & 2.00E-05 \\
        decoder\_lr & 8.00E-05 & 6.00E-05 & 6.00E-05 & 6.00E-05 \\
        mesh\_lr & 6 & 6 & 6 & 6 \\
        n\_mesh\_iters & 4 & 4 & 4 & 4 \\
        num\_iterations & 6 & 3 & 3 & 3 \\
        slot\_dropout & 0.2 & 0.2 & 0.2 & 0.2 \\
        max\_grad\_norm & 2.5 & 2.5 & 2.5 & 2.5 \\
        weight\_decay & 1.00E-05 & 1.00E-05 & 1.00E-05 & 1.00E-05 \\
        lr\_decay & 0.01 & 0.01 & 0.01 & 0.01 \\
        warm-up\_rate & 0.1 & 0.1 & 0.1 & 0.1 \\
        \bottomrule
    \end{tabular}
    \caption{Parameter settings for NYT and WebNLG experiments}
    \label{tab:exp_config}
\end{table*}
\clearpage

\twocolumn
\section{Prompting Strategy for Relational Triple Extraction}
\label{app:prompt}
We prompt the model as a relation-triple extractor with the following instructions:

\begin{enumerate}
    \item Given a sentence, extract all relational triples present in it.
    \item Each triple must be formatted as a Python dictionary with exactly three keys: "subject", "relation", and "object".
    \item The "relation" field must take its value from a fixed inventory of 24 (NYT) / 218 (WebNLG)* predefined relation types, reflecting the target ontology for our task.
    
    \textit{*Different from our data statistics because we included all the relations.}
    
    \item The model is instructed to output a Python-style list of these dictionaries.

\end{enumerate}

An example of the prompt template is as follows:

```
You are a relation-triple extractor. Given the following sentence, extract all relational triples.
Each triple must be a dict with exactly three keys: 'subject', 'relation', 'object'.
The 'relation' value must be ONE of the following 24 options:
'industry', 'place\_lived', 'founders', 'advisors', 'children', 'neighborhood\_of', 
'people', 'profession', 'place\_of\_birth', 'ethnicity', 'teams', 'company', 
'administrative\_divisions', 'place\_of\_death', 'geographic\_distribution', 
'major\_shareholders', 'nationality', 'place\_founded', 'capital', 'country', 
'religion', 'major\_shareholder\_of', 'contains', 'location'

Sentence: """[INPUT TEXT HERE]"""
Output ONLY a Python list of dictionaries. For example:
[\{'subject':'Bobby Fischer', 'relation':'nationality', 'object':'Iceland'\}, 
 \{'subject':'Iceland', 'relation':'capital', 'object':'Reykjavik'\}]
'''

The model's response is then post-processed by parsing the output using Python’s `ast.literal\_eval` to safely convert the textual list of dictionaries into a Python object. If the parsing fails, typically due to formatting errors or hallucinations, the prediction is discarded and no triples are extracted for that sample. 

In our benchmark evaluations, we did not encounter any parsing failures. However, preliminary experiments with smaller models (i.e., those with fewer than 7 billion parameters) often resulted in hallucinated or malformed outputs that could not be parsed. While these results were not formally included in this study, they highlight the importance of model scale and instruction-following capabilities in zero-shot information extraction tasks.

\section{Licensing and Terms of Use}
\subsection{Dataset}
We utilize the NYT dataset, which originates from the New York Times corpus and is distributed under the Linguistic Data Consortium (LDC) license. Additionally, we employ the WebNLG dataset, which is publicly available under the GNU General Public License v3.0 (GPL-3.0). Both datasets serve as standard benchmarks for evaluating relational extraction tasks and are also accessible through existing research repositories.

\subsection{Model and Code Release}
We will release the code after acceptance under MIT license. 

\subsection{Software Dependencies and Implementation}
The model is implemented using PyTorch\footnote{\url{https://pytorch.org/}}. We utilize the BERT-Base Cased transformer model from the Hugging Face library\footnote{\url{https://huggingface.co/google-bert/bert-base-cased}}. We use Weights \& Biases (WandB)\footnote{\url{https://wandb.ai/}} for experiment logging. We adopt the relational triple extraction evaluation metrics and preprocessing steps from an existing implementation\footnote{\url{https://github.com/DianboWork/SPN4RE}} to ensure consistency with prior work.

\subsection{Ethical and Legal Considerations}
No personally identifiable information (PII) is contained in the datasets we use. We adhere to the terms of service for all datasets and do not scrape or collect additional proprietary data.

\subsection{Information About Use Of AI Assistants}
We used ChatGPT\footnote{\url{https://chatgpt.com/}} for minor writing refinements and code debugging but ensured all final content was reviewed and verified by the authors.

\end{document}